\documentclass{article} 
\usepackage[preprint]{neurips_2023}

\usepackage{hyperref}       
\usepackage{url}            
\usepackage{booktabs}       
\usepackage{amsfonts}       
\usepackage{nicefrac}       
\usepackage{microtype}      
\usepackage{xcolor}         

\usepackage{microtype}
\usepackage{graphicx}
\usepackage{subcaption}
\usepackage{booktabs} 
\usepackage{url}
\usepackage{wrapfig}
\usepackage[normalem]{ulem}

\usepackage{hyperref}

\usepackage{amsmath}
\usepackage{amssymb}
\usepackage{mathtools}
\usepackage{amsthm}
\usepackage{multirow}
\usepackage{makecell}

\usepackage{algorithm}
\usepackage{algorithmic}

\usepackage[capitalize,noabbrev]{cleveref}

\theoremstyle{plain}
\newtheorem{theorem}{Theorem}[section]

\theoremstyle{definition}

\theoremstyle{remark}

\title{Bayesian Domain Invariant Learning via\\ Posterior Generalization of Parameter Distributions}

\author{%
Shiyu Shen\\
School of Statistics and Data Science\\
  Nankai University\\
  \texttt{shenshiyu@mail.nankai.edu.cn} \\
  \And
  Bin Pan \thanks{corresponding author} \\
  School of Statistics and Data Science \\
  Nankai University \\
  \texttt{panbin@nankai.edu.cn} \\
  \AND
  Tianyang Shi \\
  ByteDance \\
  \texttt{tirionshi@bytedance.com} \\
  \And
  Tao Li \\
  College of Computer Science \\
  Nankai University \\
  \texttt{litao@nankai.edu.cn} \\
  \And
  Zhenwei Shi \\
  College of Computer Science \\
  Beihang University \\
  \texttt{shizhenwei@buaa.edu.cn} \\
}

\begin{document}

\maketitle

\begin{abstract}
    Domain invariant learning aims to learn models that extract invariant features over various training domains, resulting in better generalization to unseen target domains. Recently, Bayesian Neural Networks have achieved promising results in domain invariant learning, but most works concentrate on aligning features distributions rather than parameter distributions. Inspired by the principle of Bayesian Neural Network, we attempt to directly learn the domain invariant posterior distribution of network parameters. We first propose a theorem to show that the invariant posterior of parameters can be implicitly inferred by aggregating posteriors on different training domains. Our assumption is more relaxed and allows us to extract more domain invariant information. We also propose a simple yet effective method, named PosTerior Generalization (PTG), that can be used to estimate the invariant parameter distribution. PTG fully exploits variational inference to approximate parameter distributions, including the invariant posterior and the posteriors on training domains. Furthermore, we develop a lite version of PTG for widespread applications. PTG shows competitive performance on various domain generalization benchmarks on DomainBed. Additionally, PTG can use any existing domain generalization methods as its prior, and combined with previous state-of-the-art method the performance can be further improved. Code will be made public.
\end{abstract}

\section{Introduction}

Distribution shift is a fundamental yet challenging problem for machine learning \citep{quinonero2008dataset,muandet2013domain}. The common assumption of independent and identically distributed data is essential for applying the networks learned from training data to test data. However, this assumption may not hold in real-world scenarios. For example, a self-driving system may be invalid in remote districts\citep{li2018rethinking,liang2018cirl}. Therefore, it's a hot topic that how to generalize a model to out-of-distribution test datasets.

Domain generalization (DG) is a solution to distribution shift \citep{zhou2021domain,gulrajani2020search}. DG usually take several training domains to train a model that generalize well on unseen test domains \citep{mixstyle,li2018domain}. One of the mainstream research interests in DG is Domain invariant learning (DIL)\citep{muandet2013domain,ilse2020diva,nguyen2021domain}. Since deep neural networks (DNN) are usually trained in an end-to-end, black-box liked way, they may fail to distinguish between informative features and unrelated features. For example, in Colored MNIST recognition task \citep{arjovsky2019invariant}, DNNs may classify digits by color rather than shape. DIL aims to extract invariant features that shared by different domains, so the disturbance from domain specific background features will be reduced. Since domain invariant features may contain more valuable information, DIL is widely acknowledged as an effective DG method.

Uncertainty is also an important consideration for out-of-distribution generalization \citep{li2022uncertainty,qiao2021uncertainty,upadhyay2021uncertainty}. Traditional DNNs are usually optimized by maximum likelihood estimation, which ignores model uncertainty and data uncertainty. Researches have validated that common DNNs are overconfident in their predictions, especially for out-of-distribution data \citep{guo2017calibration,hein2019relu,daxberger2019bayesian}. Bayesian neural network (BNN) is a well-studied approach that good at uncertainty estimation \citep{blundell2015weight,jospin2022hands,kristiadi2020being}. BNN aims to learn the posterior distributions of parameters to represent uncertainty. Some recent works have applied BNN in DG. \citet{xiao2021bit} estimate domain invariant features and classifiers by BNN, and minimize the distributional discrepancy across different domains. \citet{liu2021domain} propose a novel variational Bayesian inference framework to enforce the conditional distribution alignment via the prior distribution matching in a latent space, which also takes the marginal label shift into consideration with posterior alignment.

However, in most Bayesian domain generalization methods, BNNs are treated as a tool rather than being fully explored from the perspective of their principle: the posterior distribution of parameters. DIL learn domain invariant features by adversarial learning \citep{li2018deep,shao2019multi,li2018domain}, direct alignment \citep{li2020domain,xiao2021bit} or other methods. From the perspective of Bayes, these methods indirectly change the estimate of parameters from Maximum a Posteriori (MAP) estimate given full training data distributions to MAP given domain invariant features, which we call domain invariant parameters. Inspired by this perception, we want to directly infer the posterior distribution of domain invariant parameters from complete given domains.

In this work, we propose a novel approach to obtain the posterior of domain invariant parameters, PosTerior Generalization (PTG). PTG aggregates the posterior of parameters on different training domains to directly infer the posterior given domain invariant information. Different from other DIL methods, PTG does not need to represent domain invariant information by feature distributions. To be specific, we just assume that there exists two abstract sufficient statistics: domain invariant information $\mathcal{D}^c$ and domain specific information $\mathcal{D}^v$. $\mathcal{D}^c$ and $\mathcal{D}^v$ represent all the domain invariant information and the rest information from $\mathcal{D}$, and they should be independent. With this condition, we can directly calculate the distribution of parameter posteriors given $\mathcal{D}^c$ by Bayes formula and other formulas. Given different training domains, we can treat these domains as samples and empirically approximate the specific form of posteriors given $\mathcal{D}^c$. At last, we simplify the distribution of parameters by variational inference for easy practical application.

We also give insights into PTG from the view of feature learning. Compared with simple DIL, PTG try to make predictions by domain invariant information extract from both invariant features and part of specific features. We also provide a  lightweight, DNN based version PTG-Lite for further simplification. PTG can work as a post process that identifies the domain invariant parameters in its prior model and further aggregate the domain specific parameters, where the prior can be a model obtained by any DG method. We empirically evaluate PTG on DomainBed \citep{gulrajani2020search}. Experiments show that PTG can bring improvements across various benchmarks. Combined with the state-of-the-art competitor\citep{ref20}, PTG can further improve its performance.

Our contributions can be summarized as follows:

\begin{itemize}
    \item We introduce the analysis of parameter posterior distributions into domain generalization for the first time.
    \item Based on a relaxed assumption, we propose theories to infer the invariant posteriors, which allow us to extract more domain invariant information.
    \item We propose two simple yet effective domain generalization methods named Posterior Generalization based on our theories.
    \item Posterior Generalization achieves state-of-the-art performance on various benchmarks, and combined with other methods the performance can be further improved.
\end{itemize}

\section{Related Work}

\subsection{Domain Generalization}

Domain generalization aims to learn a generalized model by given training domains that can be applied to any unseen test domains \citep{blanchard2011generalizing,zhou2021domain,gulrajani2020search,wang2022generalizing}. There are some DG works that require only single training domain \citep{wang2021learning,qiao2020learning,gao2022loss}, but the use of multi training domains is still the mainstream setup \citep{segu2023batch,WANG2023108987,li2022invariant}. One basic DG approach is empirical risk minimization (ERM), which simply minimizes the sum of empirical risks across all domains \citep{vapnik1991principles}. \citet{gulrajani2020search} have shown that under a fair evaluation protocol, DomainBed, ERM can surprisingly outperform many DG methods. Other approaches include domain invariant learning \citep{nguyen2021domain,muandet2013domain,rame2022fishr}, data augmentation \citep{zhang2017mixup,zhang2019unseen,kang2022style}, invariant risk minimization \citep{zhou2022sparse,lin2022bayesian,arjovsky2019invariant}, meta learning \citep{li2018learning,shu2021open} and other methods \citep{hu2018does,zhang2022towards,rosenfeld2022online}.

\subsection{Domain Invariant Learning}

Domain invariant learning (DIL) is widely studied in various tasks. For example, in domain adaption \citep{csurka2017domain}, where test data without labels are available, DIL aims to learn features that shared by both training and test domains \citep{zhao2019learning}. There are theoretical guarantees that the invariant features work well on test domains \citep{ben2010theory}. However, in DG, test domains are unavailable, so DIL only learns invariant features shared by training domains. \citet{muandet2013domain} propose domain-invariant component analysis to learn an invariant transformation by minimizing the dissimilarity across domains. \citet{zhao2020domain} propose an entropy regularization term to learn conditional-invariant features across all source domains. \citet{rame2022fishr} introduce a regularization that enforces domain invariance in the space of the gradients of the loss.

\subsection{Bayesian Neural Network}
Bayesian neural network aims to estimate the uncertainty of parameters \citep{blundell2015weight,kristiadi2020being,jospin2022hands}. The key idea of BNN is to estimate the posterior distributions of parameters given training data. Recently, researches have proposed several realization methods for BNN, including Variational Inference\citep{blundell2015weight}, Markov chain Monte Carlo \citep{li2016preconditioned} and Laplace Approximate \citep{NEURIPS2021_a7c95857,pmlr-v161-kristiadi21a}. There are also modern works that apply BNN in DG. \citet{xiao2021bit} estimate the distribution of domain invariant features and classifiers and by BNN. \citet{liu2021domain} propose a variational Bayesian inference framework to enforce the conditional distribution alignment and marginal label shift alignment by distribution alignment. However, most works use BNN to estimate the distributions of features or classifiers across different domains, rather than adapting BNN from the view of parameter distributions.

\subsection{Variational Inference}
Variational inference is a popular approach to train BNNs. It approximates the true posteriors by some common distributions, such as Gaussian distribution. The distance between variational distribution and the true posterior is quantified by Kullback-Leibler (KL) divergence. \citet{blundell2015weight} propose a backpropagation-compatible algorithm for variational BNN training. \citet{kristiadi2020being} find it sufficient to build a ReLU network with a single Bayesian layer. Krishnan et~al. propose a method to choose informed weight priors in BNN by DNN.

\section{Proposed Method}

In this section, we introduce the theory of PTG and how it works. We first give some necessary notations and claims in \cref{Preliminaries}. Then, we explain the theory in \cref{bayesian_view}. The algorithm implementations of PTG are shown in \ref{full_version} and \cref{lite_version}. At last, We explain how PTG extract domain invariant information from the view of feature learning.

\subsection{Preliminaries\label{Preliminaries}}

We introduce notations for our discussions. We denote an arbitrary domain by $\mathcal{D}$, and use $\{\mathcal{D}_i\}_{i=1}^N$ to represent training domains, where $N$ is the number of training domains. For easy description in the following passage, we define $\mathcal{D}$ to be the random variable that follows the joint distribution of data $X$ and labels $Y$ in a dataset \citep{zhou2021domain}, rather than a mark of domain labels or a collection of samples. We denote parameters by $\omega$. To simplify the description, we use $p(\cdot)$ to denote the distribution of corresponding variables. For example, $p(\mathcal{D})$ means the distribution of $\mathcal{D}$.

We assume that there exist two independent sufficient statistics of each domain: domain invariant information $\mathcal{D}^c$ and domain specific information $\mathcal{D}^v$. $p(\mathcal{D}^c)$ remains constant as $\mathcal{D}$ changes, but $p(\mathcal{D}^v)$ will vary. The principle  behind this assumption is shown in \cref{principle}. We denote the domain specific information of each training domain as $\{\mathcal{D}_i^v\}_{i=1}^N$. \textbf{We do not need to assume the form of these two statistics}, while they usually exist as domain invariant and variant features \citep{shankar2018generalizing}. Furthermore, \textbf{we do not need to specify how $\mathcal{D}^c$ and $\mathcal{D}^v$ are extracted from $\mathcal{D}$}. We can approximate the posterior distribution of parameters given $\mathcal{D}^c$, $p(\omega|\mathcal{D}^c)$, even without access to $\mathcal{D}^c$.

At last, we briefly introduce how to infer the posterior of parameters by variational inference \citep{blundell2015weight}. $p(\omega|\mathcal{D}_i)$ denotes the posterior distribution of parameters given domain $\mathcal{D}_i$, and $q(\omega|\theta_i)$ denotes a variational distribution, where $\theta_i$ is the parameter of the variational distribution. We use Gaussian distribution as the variational distribution. If we train a BNN on $\mathcal{D}_i$, its loss function is:
\begin{align}
    \mathbb{D}_{KL}[q(\omega|\theta_i)||p(\omega|\mathcal{D}_i)]
    =\int q(\omega|\theta_i)log(\frac{q(\omega|\theta_i)}{p(\omega|\mathcal{D}_i)})\,d\omega.
\end{align}
By simplification, the loss function is:
\begin{align}
    \label{formula2}
    \mathbb{D}_{KL}[q(\omega|\theta_i)||p(\omega)]-\mathbb{E}_{q(\omega|\theta_i)}[log(p(\mathcal{D}_i|\omega))],
\end{align}
where $p(\omega)$ means the prior distribution of parameter, which is usually set to be standard Gaussian distribution. The first loss term can be seen as a regularization and the second term is the original negative log-likehood. In practice, the second term can be empirically optimized and the first term has an explicit expression. After training, we can approximate the intractable posterior $p(\omega|\mathcal{D}_i)$ by tractable variational distribution $q(\omega|\theta_i)$.

\subsection{Bayesian principle of PTG\label{bayesian_view}}

\begin{figure*}[htb]
    \centering
    \subcaptionbox{Bayesian view\label{vis_bayes}}[0.19\linewidth]{
        \includegraphics[width=1\linewidth]{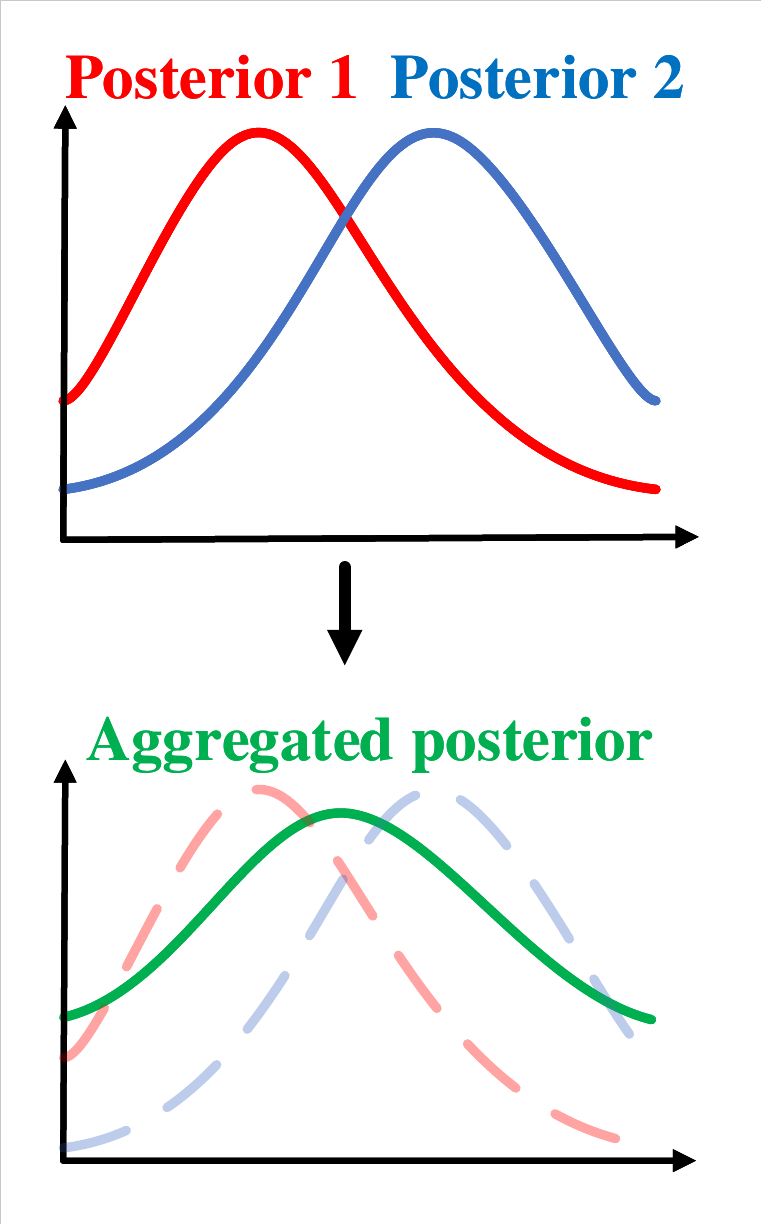}
    }
    \subcaptionbox{Feature learning view\label{vis_freq}}[0.79\linewidth]{
        \includegraphics[width=1\linewidth]{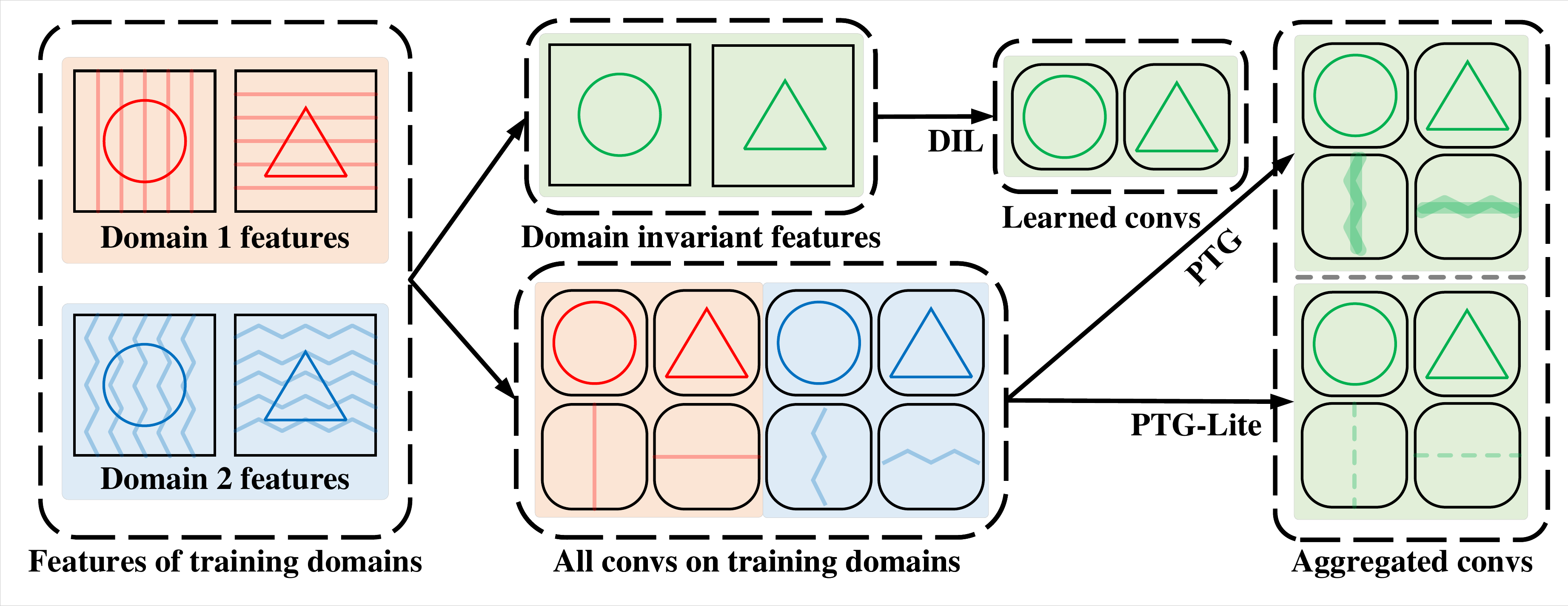}
    }
    \caption{Illustration of PTG from Bayesian view and feature learning view. From Bayesian view, PTG aggregates posteriors on each domain to infer domain invariant posteriors. From feature learning view, PTG extracts more domain invariant information from feature. DIL aims to extract invariant features while ignoring the similar but variant features. PTG methods aim to infer the invariant parameter posteriors by different aggregation approaches (separated by gray dashed line). As a result, PTG methods can preserve the invariant information from specific features.}
    \label{visualization}
\end{figure*}

To train a network that can generalize on any domain, we aim to estimate the posterior of parameters given domain invariant information $p(\omega|\mathcal{D}^c)$. However, due to the unknown content of $\mathcal{D}^c$, $p(\omega|\mathcal{D}^c)$ is intractable, let alone estimation. In fact, $\mathcal{D}^c$ and $\mathcal{D}^v$ are independent, but they always exist together. We can only get $p(\omega|\mathcal{D}^c,\mathcal{D}^c)$. Nevertheless, we can infer $p(\omega|\mathcal{D}^c )$ by the following formula:
\begin{theorem}\label{th1}
    If $\mathcal{D}^c$ and $\mathcal{D}^v$ are independent, then $p(\omega|\mathcal{D}^c)=\mathbb{E}_{p(\mathcal{D}^v)}[p(\omega|\mathcal{D}^c,\mathcal{D}^v)]$
\end{theorem}

The proof is show in \cref{proof_th1}. As a result, we can empirically estimate $p(\omega|\mathcal{D}^c)$ by sampling from $p(\mathcal{D}^v)$. Since $p(\mathcal{D}^c)$ is constant, sampling from $p(\mathcal{D}^v)$ is the same as sampling from $p(\mathcal{D})$, which is exactly $\{\mathcal{D}_i\}_{i=1}^N$. Meanwhile, $p(\omega|\mathcal{D}^c,\mathcal{D}_i^v) = p(\omega|\mathcal{D}_i)$ because $\mathcal{D}^c$ and  $\mathcal{D}_i^v$ are sufficient statistics of $\mathcal{D}_i$. Considering that $p(\omega|\mathcal{D}_i)$ can be approximate by  $q(\omega|\theta_i)$ via variational inference, we can approximate $p(\omega|\mathcal{D}^c)$ by:
\begin{align}
    p(\omega|\mathcal{D}^c)\approx \frac{\sum_{i = 1}^{N} q(\omega|\theta_i)}{N}.
\end{align}

Note that \textbf{it's the mean of distributions} $q(\omega|\theta_i)$, rather than the mean of parameters $\omega$. For the convenience of realization, we keep approximating $p(\omega|\mathcal{D}^c)$ by Gaussian variational inference. The approximate expectation and variance of $p(\omega|\mathcal{D}^c)$ can be calculated by \cref{mu_sigma}. Therefore, we replace the true domain invariant posterior by $q(w|\theta_0)$:
\begin{align}\label{formula4}
     & q(w|\theta_0) = \mathcal{N}(\mu,\sigma^2)                                                                                                                                       \\
     & \mu = \frac{\sum_{i = 1}^{N} \mathbb{E}[q(\omega|\theta_i)]}{N}                                                                                                                 \\
     & \sigma^2 = \frac{\sum_{i = 1}^{N}\mathbb{VAR}[p(\omega|\theta_i)]}{N}+\frac{\sum_{i = 1}^{N} \mathbb{E}[q(\omega|\theta_i)]^2}{N}-(\frac{\sum_{i = 1}^{N} \mathbb{E}[q(\omega|\theta_i)]}{N})^2
\end{align}
where $\mu$ and $\sigma^2$ are approximate expectation and variance. We give an illustration for the Bayesian view of PTG in \cref{vis_bayes}.

\subsection{Implementation of PTG\label{full_version}}

Although we have made some simplifications in \ref{bayesian_view} to put the theory into practice, there are still many difficulties. The first problem is the \textbf{disordered dimensions of parameters}. For example, if we train two BNNs on two domains by the same method, there's no guarantee that parameters at the same position have the same function. The first convolution kernel in the first BNN mat extract foreground features and the second convolution kernel extracts background features. The opposite situation may exist in the second BNN. If we directly calculate $p(\omega|\mathcal{D}^c)$ by PTG without addressing this issue, the aggregated convolution kernels will have great variances, and their function can be hardly explained. To mitigate this problem, we should initialize the BNN on each domain by the same, well-generalized model, e.g. a BNN trained by ERM. In this way, the function of each parameter can be approximately settled, which avoids the problem of disorder to some extent.

Another problem is the \textbf{ambiguity of classifier}. Since different training domain contains different features, the distribution of classifier, i.e. the last layers in a network, may differ a lot across domains. Similarly, if we directly calculate the posterior of domain invariant classifier, some parts of the final classifier may have large variances, which can influence the interpretability or even hurt the prediction performance. Therefore, we only construct one classifier shared by different domains, and further optimize it after the aggregation of featurizers. Besides, we design the classifier to be deterministic layers for less ambiguity.

The last problem is the \textbf{dimension reorder of parameters}. Although initialization can set parameters near extreme points, if the learning rate is too large, parameters may deviate from their local minima during training, leading to the problem of disordered dimension again. As a result, the learning rate of PTG should be carefully decayed by a rate $\alpha$, such as 0.01 times the learning rate of initialization methods. To make sure the aggregated parameters can still extract meaningful features, we further update them by ERM. The algorithm of PTG is summarized as \cref{algorithm}

\begin{algorithm}[htb]
    \small
    \caption{PTG\label{algorithm}}
    \begin{algorithmic}
        \STATE {\bfseries Input:} training domains $\{\mathcal{D}\}_{i=1}^N$
        \STATE Initialize BNN featurizers $\{f_i(\cdot)\}_{i=0}^N$ and DNN classifier $f_{cls}(\cdot)$ by a DG method
        \FOR{training iterations}
        \FOR{i=1; i$\leqslant $N; i++}
        \STATE sample minibatch data $(x_i,y_i)$ from $\mathcal{D}_i$
        \STATE calculate loss by $(f_{cls}(f_i(x_i)),y_i)$ and \cref{formula2}
        \STATE update $f_i(\cdot)$ with $\alpha$ decayed learning rate
        \ENDFOR
        \STATE update $f_0(\cdot)$ by \cref{formula4}
        \STATE merge $\{(x_i,y_i)\}_{i=1}^N$ to form $(X,Y)$
        \STATE calculate loss by $(f_{cls}(f_0(X)),Y)$ and \cref{formula2}
        \STATE update $f_0(\cdot)$ and $f_{cls}(\cdot)$ with $\alpha$ decayed learning rate
        \ENDFOR
        \STATE {\bfseries Output:} generalized network $f_{cls}(f_0(\cdot))$
    \end{algorithmic}
\end{algorithm}

\subsection{PTG-Lite\label{lite_version}}

Although PTG exploit variational inference to simplify the aggregation of posteriors, the training of BNNs and the inference of PTG are still complicated. Therefore, we further simplify PTG and propose the DNN based PTG-lite. PTG-Lite shares the same Bayesian theory with PTG, but PTG-Lite uses MAP to simplify the invariant variational distribution $q(\omega|\theta_0)$. Since we choose Gaussian distribution to be the variational distribution in PTG, the MAP estimate is exactly the expectation, so the aggregated parameters can be calculated by:
\begin{align}
    \theta_0 = \frac{\sum_{i = 1}^{N} \mathbb{E}[q(\omega|\theta_i)]}{N}.
\end{align}

Similarly, the expectations of variational distributions on different domains $q(\omega|\theta_i) $are exactly their MAP estimates. According to \cref{formula2}, the MAP estimate can be obtained by a maximum likelihood estimate (right) plus L2 regularization (left).

Different from PTG, PTG-Lite can't represent the uncertainty of parameters, so the domain specific parameters are not effectively aggregated or may even ruin the whole network. We study by experiments that it works better to drop out domain specific parameters than to replace them by mean values. We judge whether a parameter is domain specific by its coefficient of variation: if the coefficient of a parameter on different domains is greater than a given rate $\beta$, such as 0.1, we drop out this parameter. The algorithm of PTG-Lite is summarized as \cref{algorithm_lite}

\begin{algorithm}[htb]
    \small
    \caption{PTG-Lite\label{algorithm_lite}}
    \begin{algorithmic}
        \STATE {\bfseries Input:} training domains $\{\mathcal{D}\}_{i=1}^N$
        \STATE Initialize DNN featurizers $\{f_i(\cdot)\}_{i=0}^N$ and DNN classifier $f_{cls}(\cdot)$ by a DG method
        \FOR{training iterations}
        \FOR{i=1; i$\leqslant $N; i++}
        \STATE sample minibatch data $(x_i,y_i)$ from $\mathcal{D}_i$
        \STATE calculate loss by $(f_{cls}(f_i(x_i)),y_i)$ and \cref{formula2}
        \STATE update $f_i(\cdot)$ with $\alpha$ decayed learning rate
        \ENDFOR
        \STATE update $f_0(\cdot)$ by $\frac{\sum_{i = 1}^{N} f_i(\cdot) }{N}$
        \STATE drop out $f_0(\cdot)$ by coefficient of variation and rate $\beta$
        \STATE merge $\{(x_i,y_i)\}_{i=1}^N$ to form $(X,Y)$
        \STATE calculate loss by $(f_{cls}(f_0(X)),Y)$ and \cref{formula2}
        \STATE update $f_0(\cdot)$ and $f_{cls}(\cdot)$ by ERM with $\alpha$ decayed learning rate
        \ENDFOR
        \STATE {\bfseries Output:} generalized network $f_{cls}(f_0(\cdot))$
    \end{algorithmic}
\end{algorithm}

\subsection{Explanation from feature learning view\label{frequentist_view}}

\begin{figure}[htp]
    \centering
    \includegraphics[width=0.7\linewidth]{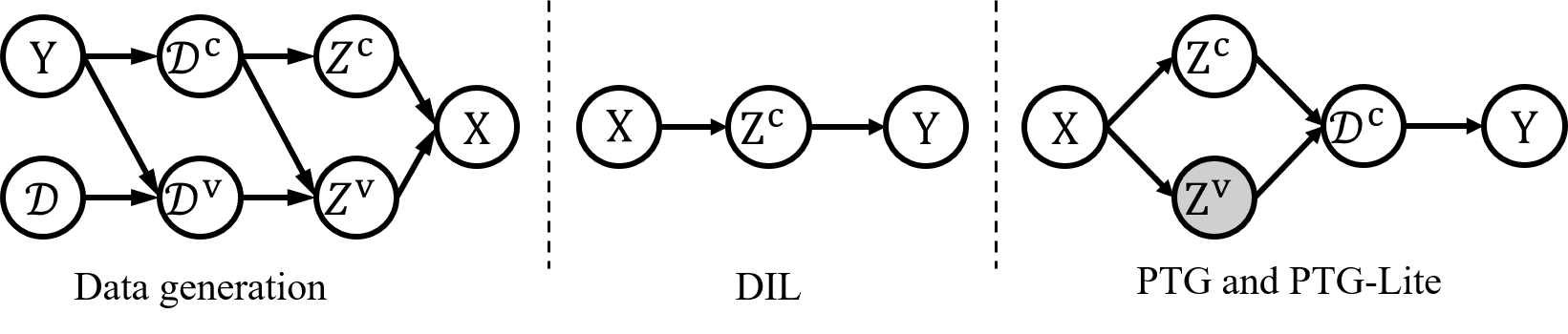}
    \caption{Casual relationships. We assume there exists domain invariant information $\mathcal{D}^c$ and domain specific information $\mathcal{D}^v$and follow the data generation assumption (left) as \citet{rosenfeld2020risks}. Most DIL (middle) makes inference by domain invariant features $Z^c$, which fail to provide enough invariant information. PTG methods (right) makes inference by domain invariant information directly, which is extracted from both invariant features and useful specific features. Gray node means the specific features are extracted by aggregated parameter posteriors.}
    \label{casual_map}
\end{figure}

Although the Bayesian principle of PTG is provided in \cref{bayesian_view}, we can give a more intuitive description of how PTG works from the view of feature learning. Moreover, the relationship between our assumption, domain invariant information, and domain invariant features can be better illustrated. As shown in \cref{casual_map}, traditional DIL makes a stronger assumption that domain invariant information exist in the form of feature maps, which may ignore some potential information that exists in specific features. In contrast, PTG directly infers the posterior distribution of parameters conditioned on the domain invariant information. And we show in the next textbf that these parameters can extract invariant information from both invariant and specific features.

There is a strong relationship between domain invariant parameters $p(\omega|\mathcal{D}^c)$ and its variation rate, details are discussed in \cref{invariant}. During the aggregation, posteriors that differ little across domains will be replaced by similar distributions; while posteriors that differ a lot will be replaced by new distributions with large variances. Consequently, PTG keeps the invariant parameters while aggregating specific parameters into more general distributions. PTG-Lite aggregates parameters by dropping out extreme specific parameters, but some specific parameters are reserved. From this perspective, PTG is more like a post process: it further identifies the remaining domain specific parameters within a prior model, and aggregate them by general parameter distributions. We give a visualization of this process in \cref{vis_freq}, where the synthetic specific features contain significant invariant information. For easy understanding, we use a whole convolution kernel to represent domain invariant or specific parameters. In fact, the domain invariant and specific parameters are mixed up.

\section{Experiments\label{experiment}}
\subsection{Experiment setup}

\textbf{Datasets.} Following \citet{gulrajani2020search}, we evaluate our method and comparison methods on four benchmarks: PACS \citep{li2017deeper}, VLCS \citep{fang2013unbiased}, OfficeHome \citep{venkateswara2017deep}, TerraIncognita \citep{beery2018recognition}.

\textbf{Evaluation protocol.} We follow the training and evaluation protocol in DomainBed. We select one domain as the target domain while the rest domains are used for training. We repeat the procedure until all domains have been used as test domains. \textbf{We select models via training domain validation set} \citep{gulrajani2020search}. The results that use other model selection methods are reported in \cref{leave1out} Each training domain is divided into 8:2 training/validation splits randomly, and the final result is selected according to the detection accuracy on these validation sets. We repeat $5\times 5$ experiments for each set up, which consist of 5 different hyperparameter samples times 5 different random seeds.

\textbf{Implementation details.} We use ResNet18 \citep{he2016deep} pre-trained on ImageNet \citep{deng2009imagenet} as the backbone networks for all models. The Results on ResNet50 are shown in \cref{result_50}. We train a BNN by other DG methods as the initializations of PTG. PTG-Lite can directly use other DG models as its initializations. All the BN layers are frozen during training. The last FC layer is replaced by a classifier with 1024 hidden units. We also apply dropout. Models are trained using the Adam optimizer. The search space of $\alpha$ is $\{0.05,0.1,0.5\}$, and $\{0.05,0.1\}$ for $\beta$. We do not use other strategies such as weight averaging \citep{cha2021swad} or ensemble learning \citep{li2023simple} to directly show the influence of PTG. More details are shown in \cref{setup_full}.

\subsection{Main Results}

We compare PTG with the following methods: Mixup \citep{yan2020improve}, CORAL \citep{ref20}, MMD \citep{ref8}, IRM \citep{arjovsky2019invariant}, GroupDRO \citep{ref34}, CAD \citep{ruan2021optimal}, VREx \citep{ref19}, SagNet \citep{ref27}, Bayes-IRM \citep{lin2022bayesian}, Fish \citep{ref37}, Fishr \citep{rame2022fishr}, ERM, ARM \citep{zhang2022towards}, SD\citep{ref29}, and SelfReg \citep{ref18}. We only compare with models that do not use large scale pre-training or ensemble learning.

\begin{table}[htb]
    \caption{\textbf{Benchmark Comparisons}. Out-of-domain classification accuracies(\%) on PACS, VLCS, OfficeHome and TerraIncognita are shown. ERM-Bayesian is a BNN \citep{blundell2015weight} trained by ERM. PTG takes ERM-Bayesian as initialization. PTG-Lite takes ERM as initialization. All models are reproduced on DomainBed. We highlight the \textbf{best}, \underline{second} and \dashuline{third} results.\label{main_result}}
    \centering
    \small
    \begin{tabular}{@{}lccccc@{}}
        \bottomrule
        \specialrule{0em}{1.5pt}{1.5pt}
        \midrule
        \textbf{Algorithm}                    & \multicolumn{1}{c}{\textbf{PACS}} & \multicolumn{1}{c}{\textbf{VLCS}} & \textbf{Office-Home}       & \textbf{TerraIncognita}    & \multicolumn{1}{c}{\textbf{Avg}} \\
        \midrule
        CAD             & 67.4 $\pm$ 6.2                    & 66.6 $\pm$ 2.2                    & 26.6 $\pm$ 9.9             & 27.5 $\pm$ 3.9             & 47.0                             \\
        IRM       & 78.9 $\pm$ 1.2                    & 73.6 $\pm$ 1.4                    & 49.7 $\pm$ 4.8             & 32.2 $\pm$ 3.4             & 58.6                             \\
        MMD                        & 80.8 $\pm$ 1.5                    & 74.2 $\pm$ 0.9                    & 58.4 $\pm$ 0.4             & 33.1 $\pm$ 9.6             & 61.6                             \\
        ARM            & 79.2 $\pm$ 0.9                    & 74.3 $\pm$ 0.9                    & 56.7 $\pm$ 0.4             & 36.6 $\pm$ 1.0             & 61.7                             \\
        GroupDRO                  & 80.3 $\pm$ 0.5                    & 73.9 $\pm$ 0.6                    & 58.0 $\pm$ 0.2             & 34.8 $\pm$ 2.2             & 61.8                             \\
        VREx                      & 81.2 $\pm$ 0.3                    & 74.4 $\pm$ 1.7                    & 59.1 $\pm$ 0.3             & 37.4 $\pm$ 0.5             & 63.0                             \\
        Bayes-IRM       & 81.1 $\pm$ 0.4                    & 74.7 $\pm$ 1.3                    & 59.3 $\pm$ 0.3             & 38.9 $\pm$ 1.1             & 63.5                             \\
        Mixup            & 79.4 $\pm$ 0.1                    & 74.4 $\pm$ 0.8                    & 60.0 $\pm$ 0.5             & \dashuline{40.3 $\pm$ 1.4} & 63.5                             \\
        Fishr             & 81.2 $\pm$ 0.9                    & 75.4 $\pm$ 0.4                    & 59.1 $\pm$ 1.1             & 40.1 $\pm$ 0.7             & 64.0                             \\
        SD                        & 80.2 $\pm$ 1.0                    & 75.0 $\pm$ 0.9                    & \textbf{62.2 $\pm$ 0.3}    & 38.6 $\pm$ 3.3             & 64.0                             \\
        SagNet                    & 81.2 $\pm$ 0.9                    & \dashuline{75.8 $\pm$ 0.4}        & 60.2 $\pm$ 1.1             & 39.3 $\pm$ 2.1             & 64.1                             \\
        SelfReg                   & \dashuline{81.8 $\pm$ 1.1}        & 75.3 $\pm$ 1.0                    & 61.2 $\pm$ 0.4             & 38.2 $\pm$ 2.4             & 64.1                             \\
        Fish                      & 80.7 $\pm$ 0.3                    & \underline{75.9 $\pm$ 0.5}        & 61.2 $\pm$ 0.4             & 39.0 $\pm$ 1.2             & 64.2                             \\
        CORAL                     & 81.2 $\pm$ 0.5                    & 75.4 $\pm$ 0.6                    & \underline{61.9 $\pm$ 0.2} & 38.7 $\pm$ 3.1             & \dashuline{64.3}                 \\
        \midrule
        ERM         & 79.8 $\pm$ 1.2                    & 75.7 $\pm$ 0.2                    & 58.9 $\pm$ 1.0             & 41.7 $\pm$ 1.5             & 64.0                             \\
        PTG-Lite                              & \underline{83.0 $\pm$ 0.3}        & \underline{75.9 $\pm$ 0.3}        & 60.9 $\pm$ 0.0             & \textbf{44.9 $\pm$ 0.4}    & \underline{66.2}                 \\
        ERM-Bayesian & 81.3 $\pm$ 0.3                    & 74.0 $\pm$ 0.7                    & 59.2 $\pm$ 0.7             & 40.9 $\pm$ 0.6             & 63.9                             \\
        PTG                                   & \textbf{83.7 $\pm$ 0.1}           & \textbf{76.1 $\pm$ 0.5}           & \dashuline{61.6 $\pm$ 0.4} & \underline{44.7 $\pm$ 1.2} & \textbf{66.5}                    \\
        \hline
        \specialrule{0em}{1.5pt}{1.5pt}
        \bottomrule
    \end{tabular}
\end{table}

The overall out-of-domain detection accuracies performances on four DG benchmarks are reported in \cref{main_result}. We show the full tables reporting the performance on each benchmark in \cref{main_result_full}. In all experiments, PTG achieves significant performance gain against ERM-Bayesian as well as the previous best results: +1.9\% in PACS, +0.2\% in VLCS, +4.4\% in TerraIncognita and +2.2\% in average compared to the previous state-of-the-art model. BNNs are recognized to have strong generalization ability because they catch uncertainty from training data. However, we observe that although ERM-Bayesian gains improvements on PACS and OfficeHome compared to ERM, the average accuracy drops, which means directly applying BNN into DG task brings little benefit. However, the outstanding performance of PTG shows that Bayesian learning is still a promising approach to solve DG problem, as long as we explore its full potential.

Besides PTG, we find that PTG-Lite also achieves good performance. PTG-Lite achieves gains against ERM by: +3.2\% in PACS, +0.2\% in VLCS, +2.0\% in OfficeHome, +3.2\% in TerraIncognita and +2.2\% in average. This may indicate that the parameters of ERM is already enough to extract necessary domain invariant features, but it also extracts some unnecessary features that may harm the generalization on target domains. Please refer to \cref{discussions} for more details.

\subsection{Combination with other methods}

\begin{table}[htb]
    \caption{\textbf{Combination with other methods.} We combine PTG with previous state-of-the-art method and report the performance on each benchmark. Each experiment is repeated 5 times.\label{combination}}
    \centering
    \small
    \begin{tabular}{@{}lcccccc@{}}
        \toprule
        \specialrule{0em}{1.5pt}{1.5pt}
        \midrule
        \textbf{Dataset}              & \textbf{Algorithm} & \multicolumn{4}{c}{\textbf{Test Domains}} & \textbf{Avg}                                            \\
        \midrule
        ~           &                    & \textbf{A}                                & \textbf{C}     & \textbf{P}     & \textbf{S}     &      \\
        \cmidrule(r){2-7}
        \multirow{4}*{PACS}                             & ERM                & 79.0 $\pm$ 0.2                            & 74.3 $\pm$ 1.7 & 94.4 $\pm$ 0.7 & 71.4 $\pm$ 2.3 & 79.8 \\
        ~                             & PTG                & 82.6 $\pm$ 0.1                            & 77.0 $\pm$ 0.3 & 94.7 $\pm$ 0.4 & 80.6 $\pm$ 0.5 & 83.7 \\
        ~                             & CORAL              & 79.6 $\pm$ 1.0                            & 75.7 $\pm$ 0.3 & 94.5 $\pm$ 0.1 & 75.2 $\pm$ 0.5 & 81.2 \\
        ~                             & CORAL-PTG          & 82.8 $\pm$ 0.7                            & 77.9 $\pm$ 0.6 & 94.9 $\pm$ 0.2 & 82.5 $\pm$ 0.3 & 84.5 \\
        \midrule
        ~           &                    & \textbf{C}                                & \textbf{L}     & \textbf{S}     & \textbf{V}     &      \\
        \cmidrule(r){2-7}
        \multirow{4}*{VLCS}                             & ERM                & 96.0 $\pm$ 0.3                            & 63.4 $\pm$ 1.1 & 70.6 $\pm$ 1.2 & 72.8 $\pm$ 1.2 & 75.7 \\
        ~                             & PTG                & 97.3 $\pm$ 0.2                            & 64.6 $\pm$ 1.2 & 68.6 $\pm$ 0.5 & 73.9 $\pm$ 0.5 & 76.1 \\
        ~                             & CORAL              & 95.3 $\pm$ 1.2                            & 64.6 $\pm$ 0.9 & 70.3 $\pm$ 0.7 & 71.4 $\pm$ 0.2 & 75.4 \\
        ~                             & CORAL-PTG          & 97.1 $\pm$ 0.6                            & 64.8 $\pm$ 1.4 & 70.4 $\pm$ 0.2 & 71.9 $\pm$ 0.8 & 76.0 \\
        \midrule
        ~     &                    & \textbf{A}                                & \textbf{C}     & \textbf{P}     & \textbf{R}     &      \\
        \cmidrule(r){2-7}
        \multirow{4}*{OfficeHome}                             & ERM                & 51.0 $\pm$ 1.6                            & 46.8 $\pm$ 1.4 & 68.3 $\pm$ 1.2 & 69.5 $\pm$ 1.5 & 58.9 \\
        ~                             & PTG                & 55.3 $\pm$ 0.5                            & 50.8 $\pm$ 0.2 & 69.7 $\pm$ 0.3 & 70.6 $\pm$ 0.4 & 61.6 \\
        ~                             & CORAL              & 55.4 $\pm$ 0.9                            & 48.7 $\pm$ 0.2 & 71.2 $\pm$ 0.6 & 72.2 $\pm$ 0.3 & 61.9 \\
        ~                             & CORAL-PTG          & 57.2 $\pm$ 1.2                            & 50.3 $\pm$ 0.8 & 71.6 $\pm$ 0.5 & 73.9 $\pm$ 0.8 & 63.3 \\
        \midrule
        ~ &                    & \textbf{L100}                             & \textbf{L38}   & \textbf{L43}   & \textbf{L46}   &      \\
        \cmidrule(r){2-7}
        \multirow{4}*{TerraIncognita}                             & ERM                & 49.5 $\pm$ 3.1                            & 32.1 $\pm$ 3.0 & 50.8 $\pm$ 0.1 & 34.2 $\pm$ 0.4 & 41.7 \\
        ~                             & PTG                & 48.6 $\pm$ 0.8                            & 40.7 $\pm$ 0.3 & 52.7 $\pm$ 0.3 & 36.8 $\pm$ 0.4 & 44.7 \\
        ~                             & CORAL              & 45.4 $\pm$ 5.2                            & 27.3 $\pm$ 6.3 & 51.4 $\pm$ 2.1 & 30.7 $\pm$ 0.9 & 38.7 \\
        ~                             & CORAL-PTG          & 46.0 $\pm$ 2.2                            & 36.1 $\pm$ 1.7 & 52.2 $\pm$ 0.7 & 33.5 $\pm$ 0.6 & 42.0 \\
        \hline
        \specialrule{0em}{1.5pt}{1.5pt}
        \bottomrule
    \end{tabular}

\end{table}

PTG needs an initialization network that trained by other DG methods. For a fair comparison, we use ERM as the initialization methods in \cref{main_result} since ERM introduces no additional DG training strategy. However, PTG can take any other DG model as its initialization, as long as the backbone structure is not changed. Here, we combine PTG with ERM and the previous state-of-the-art model CORAL to further show the power of PTG. Similarly, we initialize and further train BNNs by CORAL, and use these BNNs to initialize PTG. More combinations are shown in \cref{more_combination}

Results are presented in \cref{combination}. CORAL shows better performances than ERM with +0.3\% average out-of-domain accuracy gain. By combining PTG and CORAL, the performances are consistently improved by 2.3\% over CORAL in average. We observe that PTG can improve the accuracies across almost all experimental setups, including different prior methods, different benchmarks and different domains. We attribute this phenomenon to the dependency of the theorems of PTG and former DG methods. PTG focuses on the distribution of parameters alone, while there is no restriction about feature maps. Therefore, we believe that PTG can be easily combined with other DG methods and may get comprehensive improvements.

\section{Discussions and Limitations\label{discussions}}

\textbf{Difference between PTG and PTG-Lite.} Instead of Bayesian and non-Bayesian, the major difference between PTG and PTG-Lite roots in the aggregation process. As shown in \cref{frequentist_view}, the aggregation procedure of PTG can be regarded as making addition: we keep the domain invariant parameters while replace the domain specific parameters by general distributions. However, PTG-Lite is making subtraction: we drop the domain specific parameters directly. Both PTG and PTG-Lite can improve performance, which implies two possible research directions: (1) DG methods can benefit from some useful domain specific parameters; (2) Many DG methods already learn enough domain invariant parameters, but there are still some harmful domain specific parameters.

\textbf{PTG depends on initialization and the number of training domains.} From feature learning view, PTG is a post-procedure that refines the parameters of its prior network. Consequently, if the prior model fails to learn enough domain invariant parameters, PTG also fails. Besides, PTG estimates the invariant posterior empirically, so the number of training domain can influence the estimation reliability. We recommend the number of training domain to be 3 at least. However, we find in \cref{leave1out} that even if trained by only 2 training domains, PTG is still competitive.

\textbf{PTG is not memory efficient.} Although we have made many simplifications, the parameters on different domains have to be loaded to compute the mean and variance of parameter distributions. Besides, a BNN doubles the parameter amount of a DNN. We recommend the memory to be over 24G. Meanwhile, the training procedure of BNN is also memory consuming. However, even if we sacrifice the performance to save memory, as shown in \cref{result_50} PTG is still competitive. Furthermore, PTG just needs a few iterations (50 iterations, 1.4 epochs), so the computational costs are low.
\section{Conclusion}

In this paper, we introduce the analysis of parameter posterior distributions into Domain Invariant Learning for the first time. We theoretically show how to infer the domain invariant posterior without access to the domain invariant information condition Our relaxed assumption allow us to extract more domain invariant information. We propose a new DIL method named PTG, and explained its principles form both Bayesian view and feature learning view. Furthermore, we develop a lite, non-Bayesian version of PTG for widespread applications. The extensive experiments can show the promising performance of PTG. Besides, the combination of PTG and other methods may bring comprehensive improvements. We hope that our research promotes new research directions of examining the distributions of parameters for domain generalization.

\bibliography{PTG}

\begin{thebibliography}{73}
\providecommand{\natexlab}[1]{#1}
\providecommand{\url}[1]{\texttt{#1}}
\expandafter\ifx\csname urlstyle\endcsname\relax
  \providecommand{\doi}[1]{doi: #1}\else
  \providecommand{\doi}{doi: \begingroup \urlstyle{rm}\Url}\fi

\bibitem[Arjovsky et~al.(2019)Arjovsky, Bottou, Gulrajani, and
  Lopez-Paz]{arjovsky2019invariant}
Martin Arjovsky, L{\'e}on Bottou, Ishaan Gulrajani, and David Lopez-Paz.
\newblock Invariant risk minimization.
\newblock \emph{arXiv preprint arXiv:1907.02893}, 2019.

\bibitem[Beery et~al.(2018)Beery, Van~Horn, and Perona]{beery2018recognition}
Sara Beery, Grant Van~Horn, and Pietro Perona.
\newblock Recognition in terra incognita.
\newblock In \emph{Proceedings of the European conference on computer vision
  (ECCV)}, pp.\  456--473, 2018.

\bibitem[Ben-David et~al.(2010)Ben-David, Blitzer, Crammer, Kulesza, Pereira,
  and Vaughan]{ben2010theory}
Shai Ben-David, John Blitzer, Koby Crammer, Alex Kulesza, Fernando Pereira, and
  Jennifer~Wortman Vaughan.
\newblock A theory of learning from different domains.
\newblock \emph{Machine learning}, 79\penalty0 (1):\penalty0 151--175, 2010.

\bibitem[Blanchard et~al.(2011)Blanchard, Lee, and
  Scott]{blanchard2011generalizing}
Gilles Blanchard, Gyemin Lee, and Clayton Scott.
\newblock Generalizing from several related classification tasks to a new
  unlabeled sample.
\newblock \emph{Advances in neural information processing systems}, 24, 2011.

\bibitem[Blundell et~al.(2015)Blundell, Cornebise, Kavukcuoglu, and
  Wierstra]{blundell2015weight}
Charles Blundell, Julien Cornebise, Koray Kavukcuoglu, and Daan Wierstra.
\newblock Weight uncertainty in neural network.
\newblock In \emph{International conference on machine learning}, pp.\
  1613--1622. PMLR, 2015.

\bibitem[Cha et~al.(2021)Cha, Chun, Lee, Cho, Park, Lee, and Park]{cha2021swad}
Junbum Cha, Sanghyuk Chun, Kyungjae Lee, Han-Cheol Cho, Seunghyun Park, Yunsung
  Lee, and Sungrae Park.
\newblock Swad: Domain generalization by seeking flat minima.
\newblock \emph{Advances in Neural Information Processing Systems},
  34:\penalty0 22405--22418, 2021.

\bibitem[Csurka(2017)]{csurka2017domain}
Gabriela Csurka.
\newblock Domain adaptation for visual applications: A comprehensive survey.
\newblock \emph{arXiv preprint arXiv:1702.05374}, 2017.

\bibitem[Daxberger \& Hern{\'a}ndez-Lobato(2019)Daxberger and
  Hern{\'a}ndez-Lobato]{daxberger2019bayesian}
Erik Daxberger and Jos{\'e}~Miguel Hern{\'a}ndez-Lobato.
\newblock Bayesian variational autoencoders for unsupervised
  out-of-distribution detection.
\newblock \emph{arXiv preprint arXiv:1912.05651}, 2019.

\bibitem[Daxberger et~al.(2021)Daxberger, Kristiadi, Immer, Eschenhagen, Bauer,
  and Hennig]{NEURIPS2021_a7c95857}
Erik Daxberger, Agustinus Kristiadi, Alexander Immer, Runa Eschenhagen,
  Matthias Bauer, and Philipp Hennig.
\newblock Laplace redux - effortless bayesian deep learning.
\newblock In M.~Ranzato, A.~Beygelzimer, Y.~Dauphin, P.S. Liang, and J.~Wortman
  Vaughan (eds.), \emph{Advances in Neural Information Processing Systems},
  volume~34, pp.\  20089--20103. Curran Associates, Inc., 2021.
\newblock URL
  \url{https://proceedings.neurips.cc/paper/2021/file/a7c9585703d275249f30a088cebba0ad-Paper.pdf}.

\bibitem[Deng et~al.(2009)Deng, Dong, Socher, Li, Li, and
  Fei-Fei]{deng2009imagenet}
Jia Deng, Wei Dong, Richard Socher, Li-Jia Li, Kai Li, and Li~Fei-Fei.
\newblock Imagenet: A large-scale hierarchical image database.
\newblock In \emph{2009 IEEE conference on computer vision and pattern
  recognition}, pp.\  248--255. Ieee, 2009.

\bibitem[Fang et~al.(2013)Fang, Xu, and Rockmore]{fang2013unbiased}
Chen Fang, Ye~Xu, and Daniel~N Rockmore.
\newblock Unbiased metric learning: On the utilization of multiple datasets and
  web images for softening bias.
\newblock In \emph{Proceedings of the IEEE International Conference on Computer
  Vision}, pp.\  1657--1664, 2013.

\bibitem[Gao et~al.(2022)Gao, Gouk, Yang, and Hospedales]{gao2022loss}
Boyan Gao, Henry Gouk, Yongxin Yang, and Timothy Hospedales.
\newblock Loss function learning for domain generalization by implicit
  gradient.
\newblock In \emph{International Conference on Machine Learning}, pp.\
  7002--7016. PMLR, 2022.

\bibitem[Gulrajani \& Lopez-Paz(2020)Gulrajani and
  Lopez-Paz]{gulrajani2020search}
Ishaan Gulrajani and David Lopez-Paz.
\newblock In search of lost domain generalization.
\newblock \emph{arXiv preprint arXiv:2007.01434}, 2020.

\bibitem[Guo et~al.(2017)Guo, Pleiss, Sun, and Weinberger]{guo2017calibration}
Chuan Guo, Geoff Pleiss, Yu~Sun, and Kilian~Q Weinberger.
\newblock On calibration of modern neural networks.
\newblock In \emph{International Conference on Machine Learning}, pp.\
  1321--1330. PMLR, 2017.

\bibitem[He et~al.(2016)He, Zhang, Ren, and Sun]{he2016deep}
Kaiming He, Xiangyu Zhang, Shaoqing Ren, and Jian Sun.
\newblock Deep residual learning for image recognition.
\newblock In \emph{Proceedings of the IEEE conference on computer vision and
  pattern recognition}, pp.\  770--778, 2016.

\bibitem[Hein et~al.(2019)Hein, Andriushchenko, and Bitterwolf]{hein2019relu}
Matthias Hein, Maksym Andriushchenko, and Julian Bitterwolf.
\newblock Why relu networks yield high-confidence predictions far away from the
  training data and how to mitigate the problem.
\newblock In \emph{Proceedings of the IEEE/CVF Conference on Computer Vision
  and Pattern Recognition}, pp.\  41--50, 2019.

\bibitem[Hu et~al.(2018)Hu, Niu, Sato, and Sugiyama]{hu2018does}
Weihua Hu, Gang Niu, Issei Sato, and Masashi Sugiyama.
\newblock Does distributionally robust supervised learning give robust
  classifiers?
\newblock In \emph{International Conference on Machine Learning}, pp.\
  2029--2037. PMLR, 2018.

\bibitem[Ilse et~al.(2020)Ilse, Tomczak, Louizos, and Welling]{ilse2020diva}
Maximilian Ilse, Jakub~M Tomczak, Christos Louizos, and Max Welling.
\newblock Diva: Domain invariant variational autoencoders.
\newblock In \emph{Medical Imaging with Deep Learning}, pp.\  322--348. PMLR,
  2020.

\bibitem[Jospin et~al.(2022)Jospin, Laga, Boussaid, Buntine, and
  Bennamoun]{jospin2022hands}
Laurent~Valentin Jospin, Hamid Laga, Farid Boussaid, Wray Buntine, and Mohammed
  Bennamoun.
\newblock Hands-on bayesian neural networks—a tutorial for deep learning
  users.
\newblock \emph{IEEE Computational Intelligence Magazine}, 17\penalty0
  (2):\penalty0 29--48, 2022.

\bibitem[Kang et~al.(2022)Kang, Lee, Kim, and Kwak]{kang2022style}
Juwon Kang, Sohyun Lee, Namyup Kim, and Suha Kwak.
\newblock Style neophile: Constantly seeking novel styles for domain
  generalization.
\newblock In \emph{Proceedings of the IEEE/CVF Conference on Computer Vision
  and Pattern Recognition}, pp.\  7130--7140, 2022.

\bibitem[Koyama \& Yamaguchi(2020)Koyama and Yamaguchi]{ref18}
Masanori Koyama and Shoichiro Yamaguchi.
\newblock Out-of-distribution generalization with maximal invariant predictor.
\newblock \emph{arXiv preprint arXiv:2008.01883}, 2020.

\bibitem[Krishnan et~al.(2020)Krishnan, Subedar, and
  Tickoo]{krishnan2020specifying}
Ranganath Krishnan, Mahesh Subedar, and Omesh Tickoo.
\newblock Specifying weight priors in bayesian deep neural networks with
  empirical bayes.
\newblock In \emph{Proceedings of the AAAI Conference on Artificial
  Intelligence}, volume~34, pp.\  4477--4484, 2020.
\newblock URL \url{https://ojs.aaai.org/index.php/AAAI/article/view/5875}.

\bibitem[Kristiadi et~al.(2020)Kristiadi, Hein, and Hennig]{kristiadi2020being}
Agustinus Kristiadi, Matthias Hein, and Philipp Hennig.
\newblock Being bayesian, even just a bit, fixes overconfidence in relu
  networks.
\newblock In \emph{International conference on machine learning}, pp.\
  5436--5446. PMLR, 2020.

\bibitem[Kristiadi et~al.(2021)Kristiadi, Hein, and
  Hennig]{pmlr-v161-kristiadi21a}
Agustinus Kristiadi, Matthias Hein, and Philipp Hennig.
\newblock Learnable uncertainty under laplace approximations.
\newblock In Cassio de~Campos and Marloes~H. Maathuis (eds.), \emph{Proceedings
  of the Thirty-Seventh Conference on Uncertainty in Artificial Intelligence},
  volume 161 of \emph{Proceedings of Machine Learning Research}, pp.\
  344--353. PMLR, 27--30 Jul 2021.
\newblock URL \url{https://proceedings.mlr.press/v161/kristiadi21a.html}.

\bibitem[Krueger et~al.(2021)Krueger, Caballero, Jacobsen, Zhang, Binas, Zhang,
  Le~Priol, and Courville]{ref19}
David Krueger, Ethan Caballero, Joern-Henrik Jacobsen, Amy Zhang, Jonathan
  Binas, Dinghuai Zhang, Remi Le~Priol, and Aaron Courville.
\newblock Out-of-distribution generalization via risk extrapolation (rex).
\newblock In \emph{International Conference on Machine Learning}. PMLR, 2021.

\bibitem[Li et~al.(2022{\natexlab{a}})Li, Shen, Wang, Zhu, Li, Keutzer, and
  Zhao]{li2022invariant}
Bo~Li, Yifei Shen, Yezhen Wang, Wenzhen Zhu, Dongsheng Li, Kurt Keutzer, and
  Han Zhao.
\newblock Invariant information bottleneck for domain generalization.
\newblock In \emph{Proceedings of the AAAI Conference on Artificial
  Intelligence}, volume~36, pp.\  7399--7407, 2022{\natexlab{a}}.

\bibitem[Li et~al.(2017{\natexlab{a}})Li, Chang, Cheng, Yang, and
  P{\'o}czos]{ref20}
Chun-Liang Li, Wei-Cheng Chang, Yu~Cheng, Yiming Yang, and Barnab{\'a}s
  P{\'o}czos.
\newblock Mmd gan: Towards deeper understanding of moment matching network.
\newblock \emph{Advances in neural information processing systems}, 30,
  2017{\natexlab{a}}.

\bibitem[Li et~al.(2016)Li, Chen, Carlson, and Carin]{li2016preconditioned}
Chunyuan Li, Changyou Chen, David Carlson, and Lawrence Carin.
\newblock Preconditioned stochastic gradient langevin dynamics for deep neural
  networks.
\newblock In \emph{Thirtieth AAAI Conference on Artificial Intelligence}, 2016.

\bibitem[Li et~al.(2017{\natexlab{b}})Li, Yang, Song, and
  Hospedales]{li2017deeper}
Da~Li, Yongxin Yang, Yi-Zhe Song, and Timothy~M Hospedales.
\newblock Deeper, broader and artier domain generalization.
\newblock In \emph{Proceedings of the IEEE international conference on computer
  vision}, pp.\  5542--5550, 2017{\natexlab{b}}.

\bibitem[Li et~al.(2018{\natexlab{a}})Li, Yang, Song, and
  Hospedales]{li2018learning}
Da~Li, Yongxin Yang, Yi-Zhe Song, and Timothy Hospedales.
\newblock Learning to generalize: Meta-learning for domain generalization.
\newblock In \emph{Proceedings of the AAAI conference on artificial
  intelligence}, volume~32, 2018{\natexlab{a}}.

\bibitem[Li et~al.(2018{\natexlab{b}})Li, Pan, Wang, and Kot]{li2018domain}
Haoliang Li, Sinno~Jialin Pan, Shiqi Wang, and Alex~C Kot.
\newblock Domain generalization with adversarial feature learning.
\newblock In \emph{Proceedings of the IEEE conference on computer vision and
  pattern recognition}, pp.\  5400--5409, 2018{\natexlab{b}}.

\bibitem[Li et~al.(2020)Li, Wang, Wan, Wang, Li, and Kot]{li2020domain}
Haoliang Li, YuFei Wang, Renjie Wan, Shiqi Wang, Tie-Qiang Li, and Alex Kot.
\newblock Domain generalization for medical imaging classification with
  linear-dependency regularization.
\newblock \emph{Advances in Neural Information Processing Systems},
  33:\penalty0 3118--3129, 2020.

\bibitem[Li et~al.(2022{\natexlab{b}})Li, Dai, Ge, Liu, Shan, and
  Duan]{li2022uncertainty}
Xiaotong Li, Yongxing Dai, Yixiao Ge, Jun Liu, Ying Shan, and Ling-Yu Duan.
\newblock Uncertainty modeling for out-of-distribution generalization.
\newblock \emph{arXiv preprint arXiv:2202.03958}, 2022{\natexlab{b}}.

\bibitem[Li et~al.(2018{\natexlab{c}})Li, Tian, Gong, Liu, Liu, and
  Zhang]{li2018deep}
Ya~Li, Xinmei Tian, Mingming Gong, Yajing Liu, Tongliang Liu, and Dacheng
  Zhang, Kun anli2020domaind~Tao.
\newblock Deep domain generalization via conditional invariant adversarial
  networks.
\newblock In \emph{Proceedings of the European Conference on Computer Vision
  (ECCV)}, pp.\  624--639, 2018{\natexlab{c}}.

\bibitem[Li et~al.(2018{\natexlab{d}})Li, Motoyoshi, Sasaki, Ogata, and
  Sugano]{li2018rethinking}
Zhihao Li, Toshiyuki Motoyoshi, Kazuma Sasaki, Tetsuya Ogata, and Shigeki
  Sugano.
\newblock Rethinking self-driving: Multi-task knowledge for better
  generalization and accident explanation ability.
\newblock \emph{arXiv preprint arXiv:1809.11100}, 2018{\natexlab{d}}.

\bibitem[Li et~al.(2023)Li, Ren, Jiang, Shen, Zhang, and Li]{li2023simple}
Ziyue Li, Kan Ren, Xinyang Jiang, Yifei Shen, Haipeng Zhang, and Dongsheng Li.
\newblock Simple: Specialized model-sample matching for domain generalization.
\newblock In \emph{The Eleventh International Conference on Learning
  Representations}, 2023.

\bibitem[Liang et~al.(2018)Liang, Wang, Yang, and Xing]{liang2018cirl}
Xiaodan Liang, Tairui Wang, Luona Yang, and Eric Xing.
\newblock Cirl: Controllable imitative reinforcement learning for vision-based
  self-driving.
\newblock In \emph{Proceedings of the European conference on computer vision
  (ECCV)}, pp.\  584--599, 2018.

\bibitem[Lin et~al.(2022)Lin, Dong, Wang, and Zhang]{lin2022bayesian}
Yong Lin, Hanze Dong, Hao Wang, and Tong Zhang.
\newblock Bayesian invariant risk minimization.
\newblock In \emph{Proceedings of the IEEE/CVF Conference on Computer Vision
  and Pattern Recognition}, pp.\  16021--16030, 2022.

\bibitem[Liu et~al.(2021)Liu, Hu, Jin, Han, Xing, Ouyang, Lu, Fakhri, and
  Woo]{liu2021domain}
Xiaofeng Liu, Bo~Hu, Linghao Jin, Xu~Han, Fangxu Xing, Jinsong Ouyang, Jun Lu,
  Georges~EL Fakhri, and Jonghye Woo.
\newblock Domain generalization under conditional and label shifts via
  variational bayesian inference.
\newblock \emph{arXiv preprint arXiv:2107.10931}, 2021.

\bibitem[Muandet et~al.(2013)Muandet, Balduzzi, and
  Sch{\"o}lkopf]{muandet2013domain}
Krikamol Muandet, David Balduzzi, and Bernhard Sch{\"o}lkopf.
\newblock Domain generalization via invariant feature representation.
\newblock In \emph{International Conference on Machine Learning}, pp.\  10--18.
  PMLR, 2013.

\bibitem[Nam et~al.(2021)Nam, Lee, Park, Yoon, and Yoo]{ref27}
Hyeonseob Nam, HyunJae Lee, Jongchan Park, Wonjun Yoon, and Donggeun Yoo.
\newblock Reducing domain gap by reducing style bias.
\newblock In \emph{2021 IEEE/CVF Conference on Computer Vision and Pattern
  Recognition (CVPR)}, 2021.
\newblock \doi{10.1109/CVPR46437.2021.00858}.

\bibitem[Nguyen et~al.(2021)Nguyen, Tran, Gal, and Baydin]{nguyen2021domain}
A~Tuan Nguyen, Toan Tran, Yarin Gal, and Atilim~Gunes Baydin.
\newblock Domain invariant representation learning with domain density
  transformations.
\newblock \emph{Advances in Neural Information Processing Systems},
  34:\penalty0 5264--5275, 2021.

\bibitem[Pezeshki et~al.(2021)Pezeshki, Kaba, Bengio, Courville, Precup, and
  Lajoie]{ref29}
Mohammad Pezeshki, Oumar Kaba, Yoshua Bengio, Aaron~C Courville, Doina Precup,
  and Guillaume Lajoie.
\newblock Gradient starvation: A learning proclivity in neural networks.
\newblock In \emph{Advances in Neural Information Processing Systems},
  volume~34. Curran Associates, Inc., 2021.

\bibitem[Qiao \& Peng(2021)Qiao and Peng]{qiao2021uncertainty}
Fengchun Qiao and Xi~Peng.
\newblock Uncertainty-guided model generalization to unseen domains.
\newblock In \emph{Proceedings of the IEEE/CVF Conference on Computer Vision
  and Pattern Recognition}, pp.\  6790--6800, 2021.

\bibitem[Qiao et~al.(2020)Qiao, Zhao, and Peng]{qiao2020learning}
Fengchun Qiao, Long Zhao, and Xi~Peng.
\newblock Learning to learn single domain generalization.
\newblock In \emph{Proceedings of the IEEE/CVF Conference on Computer Vision
  and Pattern Recognition}, pp.\  12556--12565, 2020.

\bibitem[Quinonero-Candela et~al.(2008)Quinonero-Candela, Sugiyama,
  Schwaighofer, and Lawrence]{quinonero2008dataset}
Joaquin Quinonero-Candela, Masashi Sugiyama, Anton Schwaighofer, and Neil~D
  Lawrence.
\newblock \emph{Dataset shift in machine learning}.
\newblock Mit Press, 2008.

\bibitem[Rame et~al.(2022)Rame, Dancette, and Cord]{rame2022fishr}
Alexandre Rame, Corentin Dancette, and Matthieu Cord.
\newblock Fishr: Invariant gradient variances for out-of-distribution
  generalization.
\newblock In \emph{International Conference on Machine Learning}, pp.\
  18347--18377. PMLR, 2022.

\bibitem[Rosenfeld et~al.(2020)Rosenfeld, Ravikumar, and
  Risteski]{rosenfeld2020risks}
Elan Rosenfeld, Pradeep~Kumar Ravikumar, and Andrej Risteski.
\newblock The risks of invariant risk minimization.
\newblock In \emph{International Conference on Learning Representations}, 2020.

\bibitem[Rosenfeld et~al.(2022)Rosenfeld, Ravikumar, and
  Risteski]{rosenfeld2022online}
Elan Rosenfeld, Pradeep Ravikumar, and Andrej Risteski.
\newblock An online learning approach to interpolation and extrapolation in
  domain generalization.
\newblock In \emph{International Conference on Artificial Intelligence and
  Statistics}, pp.\  2641--2657. PMLR, 2022.

\bibitem[Ruan et~al.(2021)Ruan, Dubois, and Maddison]{ruan2021optimal}
Yangjun Ruan, Yann Dubois, and Chris~J Maddison.
\newblock Optimal representations for covariate shift.
\newblock \emph{arXiv preprint arXiv:2201.00057}, 2021.

\bibitem[Sagawa et~al.(2019)Sagawa, Koh, Hashimoto, and Liang]{ref34}
Shiori Sagawa, Pang~Wei Koh, Tatsunori~B Hashimoto, and Percy Liang.
\newblock Distributionally robust neural networks for group shifts: On the
  importance of regularization for worst-case generalization.
\newblock \emph{arXiv preprint arXiv:1911.08731}, 2019.

\bibitem[Segu et~al.(2023)Segu, Tonioni, and Tombari]{segu2023batch}
Mattia Segu, Alessio Tonioni, and Federico Tombari.
\newblock Batch normalization embeddings for deep domain generalization.
\newblock \emph{Pattern Recognition}, 135:\penalty0 109115, 2023.

\bibitem[Shankar et~al.(2018)Shankar, Piratla, Chakrabarti, Chaudhuri, Jyothi,
  and Sarawagi]{shankar2018generalizing}
Shiv Shankar, Vihari Piratla, Soumen Chakrabarti, Siddhartha Chaudhuri, Preethi
  Jyothi, and Sunita Sarawagi.
\newblock Generalizing across domains via cross-gradient training.
\newblock \emph{arXiv preprint arXiv:1804.10745}, 2018.

\bibitem[Shao et~al.(2019)Shao, Lan, Li, and Yuen]{shao2019multi}
Rui Shao, Xiangyuan Lan, Jiawei Li, and Pong~C Yuen.
\newblock Multi-adversarial discriminative deep domain generalization for face
  presentation attack detection.
\newblock In \emph{Proceedings of the IEEE/CVF Conference on Computer Vision
  and Pattern Recognition}, pp.\  10023--10031, 2019.

\bibitem[Shi et~al.(2022)Shi, Seely, Torr, N, Hannun, Usunier, and
  Synnaeve]{ref37}
Yuge Shi, Jeffrey Seely, Philip Torr, Siddharth N, Awni Hannun, Nicolas
  Usunier, and Gabriel Synnaeve.
\newblock Gradient matching for domain generalization.
\newblock In \emph{International Conference on Learning Representations}, 2022.

\bibitem[Shu et~al.(2021)Shu, Cao, Wang, Wang, and Long]{shu2021open}
Yang Shu, Zhangjie Cao, Chenyu Wang, Jianmin Wang, and Mingsheng Long.
\newblock Open domain generalization with domain-augmented meta-learning.
\newblock In \emph{Proceedings of the IEEE/CVF Conference on Computer Vision
  and Pattern Recognition}, pp.\  9624--9633, 2021.

\bibitem[Sun \& Saenko(2016)Sun and Saenko]{ref8}
Baochen Sun and Kate Saenko.
\newblock Deep coral: Correlation alignment for deep domain adaptation.
\newblock In \emph{Computer Vision--ECCV 2016 Workshops: Amsterdam, The
  Netherlands, October 8-10 and 15-16, 2016, Proceedings, Part III 14}.
  Springer, 2016.

\bibitem[Upadhyay et~al.(2021)Upadhyay, Chen, and
  Akata]{upadhyay2021uncertainty}
Uddeshya Upadhyay, Yanbei Chen, and Zeynep Akata.
\newblock Uncertainty-aware generalized adaptive cyclegan.
\newblock \emph{arXiv preprint arXiv:2102.11747}, 2021.

\bibitem[Vapnik(1991)]{vapnik1991principles}
Vladimir Vapnik.
\newblock Principles of risk minimization for learning theory.
\newblock \emph{Advances in neural information processing systems}, 4, 1991.

\bibitem[Venkateswara et~al.(2017)Venkateswara, Eusebio, Chakraborty, and
  Panchanathan]{venkateswara2017deep}
Hemanth Venkateswara, Jose Eusebio, Shayok Chakraborty, and Sethuraman
  Panchanathan.
\newblock Deep hashing network for unsupervised domain adaptation.
\newblock In \emph{Proceedings of the IEEE conference on computer vision and
  pattern recognition}, pp.\  5018--5027, 2017.

\bibitem[Wang et~al.(2022)Wang, Lan, Liu, Ouyang, Qin, Lu, Chen, Zeng, and
  Yu]{wang2022generalizing}
Jindong Wang, Cuiling Lan, Chang Liu, Yidong Ouyang, Tao Qin, Wang Lu, Yiqiang
  Chen, Wenjun Zeng, and Philip Yu.
\newblock Generalizing to unseen domains: A survey on domain generalization.
\newblock \emph{IEEE Transactions on Knowledge and Data Engineering}, 2022.

\bibitem[Wang et~al.(2023)Wang, Qi, Shi, and Gao]{WANG2023108987}
Ruiqi Wang, Lei Qi, Yinghuan Shi, and Yang Gao.
\newblock Better pseudo-label: Joint domain-aware label and dual-classifier for
  semi-supervised domain generalization.
\newblock \emph{Pattern Recognition}, 133:\penalty0 108987, 2023.
\newblock ISSN 0031-3203.
\newblock \doi{https://doi.org/10.1016/j.patcog.2022.108987}.
\newblock URL
  \url{https://www.sciencedirect.com/science/article/pii/S0031320322004678}.

\bibitem[Wang et~al.(2021)Wang, Luo, Qiu, Huang, and
  Baktashmotlagh]{wang2021learning}
Zijian Wang, Yadan Luo, Ruihong Qiu, Zi~Huang, and Mahsa Baktashmotlagh.
\newblock Learning to diversify for single domain generalization.
\newblock In \emph{Proceedings of the IEEE/CVF International Conference on
  Computer Vision}, pp.\  834--843, 2021.

\bibitem[Xiao et~al.(2021)Xiao, Shen, Zhen, Shao, and Snoek]{xiao2021bit}
Zehao Xiao, Jiayi Shen, Xiantong Zhen, Ling Shao, and Cees Snoek.
\newblock A bit more bayesian: Domain-invariant learning with uncertainty.
\newblock In \emph{International Conference on Machine Learning}, pp.\
  11351--11361. PMLR, 2021.

\bibitem[Yan et~al.(2020)Yan, Song, Li, Zou, and Ren]{yan2020improve}
Shen Yan, Huan Song, Nanxiang Li, Lincan Zou, and Liu Ren.
\newblock Improve unsupervised domain adaptation with mixup training.
\newblock \emph{arXiv preprint arXiv:2001.00677}, 2020.

\bibitem[Zhang et~al.(2022)Zhang, Zhang, Liu, Weller, Sch{\"o}lkopf, and
  Xing]{zhang2022towards}
Hanlin Zhang, Yi-Fan Zhang, Weiyang Liu, Adrian Weller, Bernhard Sch{\"o}lkopf,
  and Eric~P Xing.
\newblock Towards principled disentanglement for domain generalization.
\newblock In \emph{Proceedings of the IEEE/CVF Conference on Computer Vision
  and Pattern Recognition}, pp.\  8024--8034, 2022.

\bibitem[Zhang et~al.(2017)Zhang, Cisse, Dauphin, and
  Lopez-Paz]{zhang2017mixup}
Hongyi Zhang, Moustapha Cisse, Yann~N Dauphin, and David Lopez-Paz.
\newblock mixup: Beyond empirical risk minimization.
\newblock \emph{arXiv preprint arXiv:1710.09412}, 2017.

\bibitem[Zhang et~al.(2019)Zhang, Wang, Yang, Sanford, Harmon, Turkbey, Roth,
  Myronenko, Xu, and Xu]{zhang2019unseen}
Ling Zhang, Xiaosong Wang, Dong Yang, Thomas Sanford, Stephanie Harmon, Baris
  Turkbey, Holger Roth, Andriy Myronenko, Daguang Xu, and Ziyue Xu.
\newblock When unseen domain generalization is unnecessary? rethinking data
  augmentation.
\newblock \emph{arXiv preprint arXiv:1906.03347}, 2019.

\bibitem[Zhao et~al.(2019)Zhao, Des~Combes, Zhang, and
  Gordon]{zhao2019learning}
Han Zhao, Remi~Tachet Des~Combes, Kun Zhang, and Geoffrey Gordon.
\newblock On learning invariant representations for domain adaptation.
\newblock In \emph{International Conference on Machine Learning}, pp.\
  7523--7532. PMLR, 2019.

\bibitem[Zhao et~al.(2020)Zhao, Gong, Liu, Fu, and Tao]{zhao2020domain}
Shanshan Zhao, Mingming Gong, Tongliang Liu, Huan Fu, and Dacheng Tao.
\newblock Domain generalization via entropy regularization.
\newblock \emph{Advances in Neural Information Processing Systems},
  33:\penalty0 16096--16107, 2020.

\bibitem[Zhou et~al.(2021{\natexlab{a}})Zhou, Liu, Qiao, Xiang, and
  Loy]{zhou2021domain}
Kaiyang Zhou, Ziwei Liu, Yu~Qiao, Tao Xiang, and Chen~Change Loy.
\newblock Domain generalization: A survey.
\newblock 2021{\natexlab{a}}.

\bibitem[Zhou et~al.(2021{\natexlab{b}})Zhou, Yang, Qiao, and Xiang]{mixstyle}
Kaiyang Zhou, Yongxin Yang, Yu~Qiao, and Tao Xiang.
\newblock Domain generalization with mixstyle.
\newblock \emph{arXiv preprint arXiv:2104.02008}, 2021{\natexlab{b}}.

\bibitem[Zhou et~al.(2022)Zhou, Lin, Zhang, and Zhang]{zhou2022sparse}
Xiao Zhou, Yong Lin, Weizhong Zhang, and Tong Zhang.
\newblock Sparse invariant risk minimization.
\newblock In \emph{International Conference on Machine Learning}, pp.\
  27222--27244. PMLR, 2022.

\end{thebibliography}


\begin{thebibliography}{3}
\providecommand{\natexlab}[1]{#1}
\providecommand{\url}[1]{\texttt{#1}}
\expandafter\ifx\csname urlstyle\endcsname\relax
  \providecommand{\doi}[1]{doi: #1}\else
  \providecommand{\doi}{doi: \begingroup \urlstyle{rm}\Url}\fi

\bibitem[Bengio \& LeCun(2007)Bengio and LeCun]{Bengio+chapter2007}
Yoshua Bengio and Yann LeCun.
\newblock Scaling learning algorithms towards {AI}.
\newblock In \emph{Large Scale Kernel Machines}. MIT Press, 2007.

\bibitem[Goodfellow et~al.(2016)Goodfellow, Bengio, Courville, and
  Bengio]{goodfellow2016deep}
Ian Goodfellow, Yoshua Bengio, Aaron Courville, and Yoshua Bengio.
\newblock \emph{Deep learning}, volume~1.
\newblock MIT Press, 2016.

\bibitem[Hinton et~al.(2006)Hinton, Osindero, and Teh]{Hinton06}
Geoffrey~E. Hinton, Simon Osindero, and Yee~Whye Teh.
\newblock A fast learning algorithm for deep belief nets.
\newblock \emph{Neural Computation}, 18:\penalty0 1527--1554, 2006.

\end{thebibliography}
\bibliographystyle{iclr2024_conference}

\newpage
\appendix
\onecolumn
\begin{huge}
    \textbf{Appednix}
\end{huge}

\section{Principle of the assumption that $\mathcal{D}^c$ and $\mathcal{D}^v$ exist and are independent.\label{principle}}

We do not assume the form of existence of $\mathcal{D}^c$ and $\mathcal{D}^v$, or how they can be obtained. They are two abstract statistics that contain all the domain invariant information and the rest information of $\mathcal{D}$. We only assume their existence and independence. Based on our assumption, the domain invariant features are part of $\mathcal{D}^c$ or can be further inferred from $\mathcal{D}^c$.

We can provide the rationality of our assumption. (1)If $\mathcal{D}^c$ didn't exist, which means there were no invariant information among domains, then Domain Generalization would have no solution. (2)If $\mathcal{D}^v$ didn't exist, which means there were only invariant information among domains, then there would be no need for further generalization. (3)If $\mathcal{D}^c$ and $\mathcal{D}^v$ exist but could not be separated, then the domain invariant features should always contain specific information (otherwise, part of $\mathcal{D}^c$ should be independent from $\mathcal{D}^v$, and we can take it as the true $\mathcal{D}^c$), and domain invariant learning would have no solution.

\section{Proof of \cref{th1}}\label{proof_th1}
If $\mathcal{D}^c and \mathcal{D}^v$ are independent, then $p(\omega|\mathcal{D}^c)=\mathbb{E}_{p(\mathcal{D}^v)}[p(\omega|\mathcal{D}^c,\mathcal{D}^v)]$
\begin{proof}
    \begin{align}
          & \mathbb{E}_{p(\mathcal{D}^v)}[p(\omega|\mathcal{D}^c,\mathcal{D}^v)]                                                \\
        = & \int p(\mathcal{D}^v)p(\omega|\mathcal{D}^c,\mathcal{D}^v) \,d\mathcal{D}^v                                         \\
        = & \int p(\mathcal{D}^v)\frac{p(\omega,\mathcal{D}^c,\mathcal{D}^v)}{p(\mathcal{D}^c,\mathcal{D}^v)} \,d\mathcal{D}^v.
    \end{align}
    Since $\mathcal{D}^c$ and  $\mathcal{D}^v$ are independent,
    \begin{align}
          & \int p(\mathcal{D}^v)\frac{p(\omega,\mathcal{D}^c,\mathcal{D}^v)}{p(\mathcal{D}^c,\mathcal{D}^v)} \,d\mathcal{D}^v   \\
        = & \int p(\mathcal{D}^v)\frac{p(\omega,\mathcal{D}^c,\mathcal{D}^v)}{p(\mathcal{D}^c)p(\mathcal{D}^v)} \,d\mathcal{D}^v \\
        = & \int \frac{p(\omega,\mathcal{D}^c,\mathcal{D}^v)}{p(\mathcal{D}^c)} \,d\mathcal{D}^v                                 \\
        = & \frac{p(\omega,\mathcal{D}^c)}{p(\mathcal{D}^c)}                                                                     \\
        = & p(\omega|\mathcal{D}^c)
    \end{align}
\end{proof}
\newpage
\section{Mathematical Derivation for the Expectation and Variance of $p(\omega|\mathcal{D})$}\label{mu_sigma}
We use $f_{p(\cdot)}$ to represent the density function of corresponding distribution.
\begin{align}
    f_{p(\omega|\mathcal{D}^c)}\approx      & \frac{\sum_{i = 1}^{N} f_{q(\omega|\theta_i)}}{N}                                                                                                                                                                        \\
    \mathbb{E}[p(\omega|\mathcal{D}^c)] =   & \int xf_{p(\omega|\mathcal{D}^c)}(x) \,dx                                                                                                                                                                                \\
    =                                       & \int x\frac{\sum_{i = 1}^{N} f_{q(\omega|\theta_i)}(x)}{N} \,dx                                                                                                                                                          \\
    \approx                                 & \frac{\sum_{i = 1}^{N} \mathbb{E}[q(\omega|\theta_i)]}{N}                                                                                                                                                                \\
    \mathbb{E}[p(\omega|\mathcal{D}^c)^2] = & \int x^2f_{p(\omega|\mathcal{D}^c)}(x) \,dx                                                                                                                                                                              \\
    =                                       & \int x^2\frac{\sum_{i = 1}^{N} f_{q(\omega|\theta_i)}(x)}{N} \,dx                                                                                                                                                        \\
    \approx                                 & \frac{\sum_{i = 1}^{N} \mathbb{E}[q(\omega|\theta_i)^2]}{N}                                                                                                                                                              \\
    \mathbb{VAR}[p(\omega|\mathcal{D}^c)] = & \mathbb{E}[p(\omega|\mathcal{D}^c)^2] - \mathbb{E}[p(\omega|\mathcal{D}^c)]^2                                                                                                                                            \\
    \approx                                 & \frac{\sum_{i = 1}^{N} \mathbb{E}[q(\omega|\theta_i)^2]}{N}-(\frac{\sum_{i = 1}^{N} \mathbb{E}[q(\omega|\theta_i)]}{N})^2                                                                                                \\
    =                                       & \frac{\sum_{i = 1}^{N} (\mathbb{E}[q(\omega|\theta_i)^2]-\mathbb{E}[q(\omega|\theta_i)]^2)}{N}+\frac{\sum_{i = 1}^{N} \mathbb{E}[q(\omega|\theta_i)]^2}{N}-(\frac{\sum_{i = 1}^{N} \mathbb{E}[q(\omega|\theta_i)]}{N})^2 \\
    =                                       & \frac{\sum_{i = 1}^{N}\mathbb{VAR}[p(\omega|\theta_i)]}{N}+\frac{\sum_{i = 1}^{N} \mathbb{E}[q(\omega|\theta_i)]^2}{N}-(\frac{\sum_{i = 1}^{N} \mathbb{E}[q(\omega|\theta_i)]}{N})^2
\end{align}

\section{Relationship between $p(\omega|\mathcal{D}^c)$and its variation rate.\label{invariant}}

Since $\mathcal{D}_i^c$ follows the same distribution among domains, $p(\omega|\mathcal{D}_i^c)$ also follows the same distribution among domains. On the other hand, $\mathcal{D}_i^v$ follows different distributions among domains, so $p(\omega|\mathcal{D}_i^v)$ also follows different distributions. From another perspective, in Bayesian Neural Networks, larger variance in a parameter implies higher uncertainty. In DG context, uncertainty mainly comes from the difference between domains, so parameters that change a lot among domains(high variance) are more likely to extract domain specific features.

\section{Experiment setup\label{setup_full}}

\textbf{Datasets.} Following Gulrajani and Lopez-Paz \citep{gulrajani2020search}, we evaluate our method and comparison methods on four benchmarks: PACS \citep{li2017deeper} containing 9,991 images of 7 classes across 4 domains \{photo, art, cartoon, sketch\}, VLCS \citep{fang2013unbiased} containing 10,729 images of 5 classes across 4 domains \{VOC2007, LabelMe, Caltech101,
SUN09\}, OfficeHome \citep{venkateswara2017deep} containing 15,588 images of 65 classes across 4 domains \{art, clipart, product, real\}, TerraIncognita \citep{beery2018recognition} containing 24,788 images of 10 classes across 4 domains \{L100, L38, L43, L46\}.

\textbf{Evaluation protocol.} For a fair comparison, we follow the training and evaluation protocol in DomainBed. We select one domain as the target domain while the rest domains are used for training. We repeat the procedure until all domains have been used as test domains. We select models via training domain validation set \citep{gulrajani2020search}. Each training domain is divided into 8:2 training/validation splits randomly, and the final result is selected according to the detection accuracy on these validation sets. We repeat $5\times 5$ experiments for each set up, which consist of 5 different hyperparameter samples times 5 different random seeds. We select the best hyperparameter and report the mean and standard deviation of test domain classification accuracies from 5 random runs.

\textbf{Implementation details.} We use ResNet18 \citep{he2016deep} pre-trained on ImageNet \citep{deng2009imagenet} as the backbone networks for all models. However, PTG and PTG-Lite need pre-trained DG models as their initializations. The initializations of PTG should be BNNs. To reduce the computation cost, we specify the prior of BNNs by DG trained DNNs \citep{krishnan2020specifying} and further train the BNN in the same way. Then, the BNNs are used as the initializations of PTG. PTG-Lite can directly use the DNNs trained by other DG methods as its initializations. All the BN layers are frozen during training. The last FC layer is replaced by a classifier with 1024 hidden units. We also apply dropout where the dropout rate is selected by DomainBed. Models are trained using the Adam optimizer. The search space of decay rate $\alpha$ is $\{0.05,0.1,0.5\}$, and the search space of coefficient of variation $\beta$ is $\{0.05,0.1\}$. Since PTG is a post-processing algorithm, we do not use any other strategies such as weight averaging \citep{cha2021swad} or ensemble learning \citep{li2023simple}, to directly show the influence of PTG.

\section{Leave-one-domain-out cross-validation Results\label{leave1out}}

We didn't report the results of models selected by leave-one-domain-out cross-validation in the main body, because we recommend that the number of training domains should be 3 at least. In \cref{experiment_leave}, we find that even if we use leave-one-domain-out cross-validation, which means we only use two training domains and one validation domain, the performance is still good enough. However, we still suggest that the number of training domains should be adequate just in case.

\begin{table}[!h]
    \caption{\textbf{Leave-one-domain-out cross-validation Results}. All models use ResNet18 as backbones.\label{experiment_leave}}
    \centering
    \begin{tabular}{@{}lccccc@{}}
        \bottomrule
        \specialrule{0em}{1.5pt}{1.5pt}
        \midrule
        \textbf{Algorithm} & \multicolumn{1}{c}{\textbf{PACS}} & \multicolumn{1}{c}{\textbf{VLCS}} & \textbf{Office-Home} & \textbf{TerraIncognita} & \multicolumn{1}{c}{\textbf{Avg}} \\
        \midrule
        SelfReg            & 83.4 $\pm$ 0.8                    & 78.9 $\pm$ 0.2                    & 66.2 $\pm$ 0.6       & 46.3 $\pm$ 1.0          & 68.7                             \\
        Fish               & 84.4 $\pm$ 1.1                    & 80.4 $\pm$ 0.4                    & 65.0 $\pm$ 0.4       & 43.9 $\pm$ 1.6          & 68.4                             \\
        CORAL              & 84.7 $\pm$ 0.7                    & 78.9 $\pm$ 0.5                    & 65.9 $\pm$ 0.4       & 45.8 $\pm$ 1.6          & 68.8                             \\
        \midrule
        ERM                & 82.7 $\pm$ 1.3                    & 77.0 $\pm$ 0.4                    & 65.5 $\pm$ 1.1       & 41.2 $\pm$ 0.9          & 66.6                             \\
        PTG-Lite           & 84.5 $\pm$ 0.3                    & 76.1 $\pm$ 0.2                    & 67.6 $\pm$ 0.2       & 47.7 $\pm$ 0.7          & 69.0                             \\
        ERM-Bayesian       & 84.7 $\pm$ 0.5                    & 76.6 $\pm$ 0.4                    & 63.8 $\pm$ 0.4       & 43.7 $\pm$ 0.9          & 67.2                             \\
        PTG                & 86.3 $\pm$ 0.4                    & 76.3 $\pm$ 0.5                    & 67.1 $\pm$ 0.3       & 46.3 $\pm$ 0.7          & 69.0                             \\
        \hline
        \specialrule{0em}{1.5pt}{1.5pt}
        \bottomrule
    \end{tabular}
\end{table}
\newpage
\section{Results on ResNet50\label{result_50}}

We didn't provide results on ResNet50 in the main body for consideration of both GPU memory cost and fairness. We develop a degraded PTG training algorithm to save memory(change the loss of average outputs into average loss of outputs), but this behavior hurts the performance. Consequently, we have reported the performance of all models based on ResNet18 for fair comparison. The performance on ResNet50 is shown in \cref{res50}, where we continue to achieve optimal performance. Again we want to remind that the results on ResNet50 can't reflect the full ability of PTG.

\begin{table}[!h]
    \caption{\textbf{Results on ResNet50}. All methods use training-domain validation set to select models. We report the performance of competitors according to their original papers. We want to remind again that we sacrifice the performance of PTG on ResNet50 to save GPU memory. For more results, please refer to DomainBed.\label{res50}}
    \centering
    \begin{tabular}{@{}lccccc@{}}
        \bottomrule
        \specialrule{0em}{1.5pt}{1.5pt}
        \midrule
        \textbf{Algorithm} & \multicolumn{1}{c}{\textbf{PACS}} & \multicolumn{1}{c}{\textbf{VLCS}} & \textbf{Office-Home} & \textbf{TerraIncognita} & \multicolumn{1}{c}{\textbf{Avg}} \\
        \midrule
        SelfReg            & 85.6 $\pm$ 0.4                    & 77.8 $\pm$ 0.9                    & 67.9 $\pm$ 0.7       & 47.0 $\pm$ 0.3          & 70.0                             \\
        Fish               & 85.5 $\pm$ 0.3                    & 77.8 $\pm$ 0.3                    & 68.6 $\pm$ 0.4       & 45.1 $\pm$ 1.3          & 69.3                             \\
        CORAL              & 86.2 $\pm$ 0.3                    & 78.8 $\pm$ 0.6                    & 68.7 $\pm$ 0.3       & 47.6 $\pm$ 1.0          & 70.3                             \\
        \midrule
        ERM                & 85.5 $\pm$ 0.2                    & 77.5 $\pm$ 0.4                    & 66.5 $\pm$ 0.3       & 46.1 $\pm$ 1.8          & 68.9                             \\
        PTG-Lite           & 87.3 $\pm$ 0.2                    & 79.6 $\pm$ 0.5                    & 70.0 $\pm$ 0.3       & 49.2 $\pm$ 0.7          & 71.5                             \\
        ERM-Bayesian       & 85.8 $\pm$ 0.5                    & 77.7 $\pm$ 0.3                    & 67.1 $\pm$ 0.2       & 45.5 $\pm$ 0.8          & 69.0                             \\
        PTG                & 86.7 $\pm$ 0.2                    & 79.4 $\pm$ 0.5                    & 69.4 $\pm$ 0.6       & 48.5 $\pm$ 1.1          & 71.0                             \\
        \hline
        \specialrule{0em}{1.5pt}{1.5pt}
        \bottomrule
    \end{tabular}
\end{table}

\section{More combinations\label{more_combination}}
PTG can be combined with most existing methods since it functions as a post-process strategy. However, demonstrating the combination of PTG with all models is unnecessary. By \cref{combination}, we already show that PTG can further promote the performance by combination. We add some experiments to show more combinations in \cref{combination_more}, which show that all the combinations can bring promotions to the original method.

\begin{table}[!h]
    \caption{\textbf{More Combinations}. All methods use training domain validation to select models. All models use ResNet18 as backbones.\label{combination_more}}
    \centering
    \begin{tabular}{@{}lccccc@{}}
        \bottomrule
        \specialrule{0em}{1.5pt}{1.5pt}
        \midrule
        \textbf{Algorithm} & \multicolumn{1}{c}{\textbf{PACS}} & \multicolumn{1}{c}{\textbf{VLCS}} & \textbf{Office-Home} & \textbf{TerraIncognita} & \multicolumn{1}{c}{\textbf{Avg}} \\
        \midrule
        ERM                & 79.8 $\pm$ 1.2                    & 75.7 $\pm$ 0.2                    & 58.9 $\pm$ 1.0       & 41.7 $\pm$ 1.5          & 64.0                             \\
        PTG                & 83.7 $\pm$ 0.1                    & 76.1 $\pm$ 0.5                    & 61.6 $\pm$ 0.4       & 44.7 $\pm$ 1.2          & 66.5                             \\
        SelfReg            & 81.8 $\pm$ 1.1                    & 75.3 $\pm$ 1.0                    & 61.2 $\pm$ 0.4       & 38.2 $\pm$ 2.4          & 64.1                             \\
        SelfReg-PTG        & 85.3 $\pm$ 0.4                    & 75.2 $\pm$ 0.4                    & 63.6 $\pm$ 0.5       & 42.6 $\pm$ 0.9          & 66.7                             \\
        Fish               & 80.7 $\pm$ 0.3                    & 75.9 $\pm$ 0.5                    & 61.2 $\pm$ 0.4       & 39.0 $\pm$ 1.2          & 64.2                             \\
        Fish-PTG           & 84.9 $\pm$ 0.2                    & 76.4 $\pm$ 0.3                    & 63.6 $\pm$ 0.4       & 43.3 $\pm$ 1.1          & 67.1                             \\
        CORAL              & 81.2 $\pm$ 0.5                    & 75.4 $\pm$ 0.6                    & 61.9 $\pm$ 0.2       & 38.7 $\pm$ 3.1          & 64.3                             \\
        CORAL-PTG          & 84.5 $\pm$ 0.4                    & 76.0 $\pm$ 0.8                    & 63.3 $\pm$ 0.8       & 42.0 $\pm$ 1.3          & 66.5                             \\
        \hline
        \specialrule{0em}{1.5pt}{1.5pt}
        \bottomrule
    \end{tabular}
\end{table}

\section{Full results of \cref{main_result}\label{main_result_full}}

\begin{table}[!h]
    \caption{\textbf{PACS Comparisons}. Out-of-domain classification accuracies(\%) on PACS are shown. ERM-Bayesian is a BNN \citep{blundell2015weight} trained by ERM. PTG takes ERM-Bayesian as initialization. PTG-Lite takes ERM as initialization. All models are reproduced on DomainBed. We highlight the \textbf{best}, \underline{second} and \dashuline{third} results.}
    \centering
    \begin{tabular}{@{}lccccc@{}}
        \bottomrule
        \specialrule{0em}{1.5pt}{1.5pt}
        \midrule
        \textbf{Algorithm}                    & \textbf{A}     & \textbf{C}     & \textbf{P}     & \textbf{S}      & \textbf{Avg}     \\
        \midrule
        CAD             & 66.9 $\pm$ 2.7 & 62.8 $\pm$ 7.9 & 82.1 $\pm$ 3.7 & 57.6 $\pm$ 10.8 & 67.4             \\
        IRM       & 75.1 $\pm$ 2.5 & 74.0 $\pm$ 0.9 & 92.9 $\pm$ 1.6 & 73.4 $\pm$ 0.6  & 78.9             \\
        MMD\                        & 82.3 $\pm$ 1.4 & 75.6 $\pm$ 0.9 & 92.8 $\pm$ 0.2 & 72.7 $\pm$ 0.5  & 80.8             \\
        ARM            & 80.9 $\pm$ 0.6 & 70.9 $\pm$ 0.5 & 91.5 $\pm$ 0.3 & 73.4 $\pm$ 1.2  & 79.2             \\
        GroupDRO                  & 79.4 $\pm$ 0.9 & 75.0 $\pm$ 0.6 & 92.7 $\pm$ 0.3 & 74.2 $\pm$ 2.0  & 80.3             \\
        VREx                      & 82.3 $\pm$ 1.6 & 75.4 $\pm$ 0.6 & 93.2 $\pm$ 0.5 & 74.0 $\pm$ 1.7  & 81.2             \\
        Bayes-IRM       & 80.9 $\pm$ 0.7 & 75.5 $\pm$ 1.3 & 93.7 $\pm$ 0.6 & 74.2 $\pm$ 1.2  & 81.1             \\
        Mixup            & 78.7 $\pm$ 1.8 & 73.0 $\pm$ 1.2 & 94.0 $\pm$ 0.3 & 71.7 $\pm$ 1.0  & 79.4             \\
        Fishr             & 84.1 $\pm$ 0.2 & 74.4 $\pm$ 0.7 & 92.7 $\pm$ 0.1 & 73.5 $\pm$ 2.0  & 81.2             \\
        SD                        & 80.4 $\pm$ 1.3 & 74.6 $\pm$ 0.5 & 92.4 $\pm$ 0.2 & 73.4 $\pm$ 1.2  & 80.2             \\
        SagNet                    & 79.6 $\pm$ 1.7 & 75.2 $\pm$ 0.8 & 93.7 $\pm$ 0.7 & 76.2 $\pm$ 0.8  & 81.2             \\
        SelfReg                   & 81.7 $\pm$ 0.8 & 75.2 $\pm$ 1.3 & 92.5 $\pm$ 0.4 & 77.8 $\pm$ 1.1  & \dashuline{81.8} \\
        Fish                      & 80.1 $\pm$ 1.2 & 73.8 $\pm$ 0.8 & 94.4 $\pm$ 0.2 & 74.5 $\pm$ 1.0  & 80.7             \\
        CORAL                     & 79.6 $\pm$ 1.0 & 75.7 $\pm$ 0.3 & 94.5 $\pm$ 0.1 & 75.2 $\pm$ 0.5  & 81.2             \\
        \midrule
        ERM         & 79.0 $\pm$ 0.2 & 74.3 $\pm$ 1.7 & 94.4 $\pm$ 0.7 & 71.4 $\pm$ 2.3  & 79.8             \\
        PTG-Lite                              & 82.4 $\pm$ 0.9 & 75.0 $\pm$ 0.6 & 94.9 $\pm$ 0.5 & 79.6 $\pm$ 0.7  & \underline{83.0} \\
        ERM-Bayesian & 79.2 $\pm$ 1.0 & 73.9 $\pm$ 0.8 & 93.6 $\pm$ 0.2 & 78.6 $\pm$ 0.9  & 81.3             \\
        PTG                                   & 82.6 $\pm$ 0.1 & 77.0 $\pm$ 0.3 & 94.7 $\pm$ 0.4 & 80.6 $\pm$ 0.5  & \textbf{83.7}    \\
        \hline
        \specialrule{0em}{1.5pt}{1.5pt}
        \bottomrule
    \end{tabular}
\end{table}

\begin{table}[!h]
    \caption{\textbf{VLCS Comparisons}. Out-of-domain classification accuracies(\%) on VLCS are shown. ERM-Bayesian is a BNN \citep{blundell2015weight} trained by ERM. PTG takes ERM-Bayesian as initialization. PTG-Lite takes ERM as initialization. All models are reproduced on DomainBed. We highlight the \textbf{best}, \underline{second} and \dashuline{third} results.}
    \centering
    \begin{tabular}{@{}lccccc@{}}
        \bottomrule
        \specialrule{0em}{1.5pt}{1.5pt}
        \midrule
        \textbf{Algorithm}                    & \textbf{C}     & \textbf{L}     & \textbf{S}     & \textbf{V}     & \textbf{Avg}     \\
        \midrule
        CAD             & 84.9 $\pm$ 5.5 & 61.3 $\pm$ 0.2 & 59.0 $\pm$ 3.1 & 61.3 $\pm$ 0.6 & 66.6             \\
        IRM       & 94.7 $\pm$ 1.6 & 62.6 $\pm$ 0.9 & 68.6 $\pm$ 1.8 & 68.7 $\pm$ 4.2 & 73.6             \\
        MMD                        & 94.4 $\pm$ 1.1 & 60.7 $\pm$ 2.1 & 69.5 $\pm$ 1.2 & 72.0 $\pm$ 4.3 & 74.2             \\
        ARM            & 95.4 $\pm$ 1.2 & 60.3 $\pm$ 1.7 & 69.0 $\pm$ 2.2 & 73.4 $\pm$ 1.6 & 74.3             \\
        GroupDRO                  & 94.5 $\pm$ 1.3 & 60.6 $\pm$ 1.9 & 66.7 $\pm$ 1.8 & 73.9 $\pm$ 1.8 & 73.9             \\
        VREx                      & 94.5 $\pm$ 1.5 & 60.5 $\pm$ 2.3 & 70.2 $\pm$ 1.4 & 72.3 $\pm$ 2.3 & 74.4             \\
        Bayes-IRM       & 94.0 $\pm$ 1.9 & 62.2 $\pm$ 2.0 & 69.7 $\pm$ 1.6 & 72.8 $\pm$ 1.9 & 74.7             \\
        Mixup            & 95.5 $\pm$ 0.3 & 61.0 $\pm$ 0.6 & 69.2 $\pm$ 1.1 & 71.7 $\pm$ 1.7 & 74.4             \\
        Fishr             & 95.9 $\pm$ 0.9 & 60.6 $\pm$ 1.5 & 68.1 $\pm$ 1.2 & 73.4 $\pm$ 1.7 & 75.4             \\
        SD                        & 94.8 $\pm$ 0.9 & 61.3 $\pm$ 1.2 & 69.2 $\pm$ 0.7 & 71.6 $\pm$ 1.2 & 75.0             \\
        SagNet                    & 95.8 $\pm$ 0.9 & 64.0 $\pm$ 0.8 & 69.6 $\pm$ 1.0 & 73.8 $\pm$ 0.9 & \dashuline{75.8} \\
        SelfReg                   & 95.4 $\pm$ 0.6 & 63.2 $\pm$ 1.2 & 68.9 $\pm$ 1.5 & 73.4 $\pm$ 0.5 & 75.3             \\
        Fish                      & 97.0 $\pm$ 0.5 & 62.3 $\pm$ 1.0 & 70.7 $\pm$ 0.9 & 73.5 $\pm$ 0.7 & \underline{75.9} \\
        CORAL                     & 95.3 $\pm$ 1.2 & 64.6 $\pm$ 0.9 & 70.3 $\pm$ 0.7 & 71.4 $\pm$ 0.2 & 75.4             \\
        \midrule
        ERM         & 96.0 $\pm$ 0.3 & 63.4 $\pm$ 1.1 & 70.6 $\pm$ 1.2 & 72.8 $\pm$ 1.2 & 75.7             \\
        PTG-Lite                              & 96.8 $\pm$ 0.2 & 63.9 $\pm$ 0.2 & 69.5 $\pm$ 0.7 & 72.9 $\pm$ 0.7 & \underline{75.9} \\
        ERM-Bayesian & 96.2 $\pm$ 0.9 & 62.2 $\pm$ 0.6 & 67.3 $\pm$ 1.0 & 70.4 $\pm$ 0.8 & 74.0             \\
        PTG                                   & 97.3 $\pm$ 0.2 & 64.6 $\pm$ 1.2 & 68.6 $\pm$ 0.5 & 73.9 $\pm$ 0.5 & \textbf{76.1}    \\
        \hline
        \specialrule{0em}{1.5pt}{1.5pt}
        \bottomrule
    \end{tabular}
\end{table}

\begin{table}[!h]
    \caption{\textbf{OfficeHome Comparisons}. Out-of-domain classification accuracies(\%) on OfficeHome are shown. ERM-Bayesian is a BNN \citep{blundell2015weight} trained by ERM. PTG takes ERM-Bayesian as initialization. PTG-Lite takes ERM as initialization. All models are reproduced on DomainBed. We highlight the \textbf{best}, \underline{second} and \dashuline{third} results.}
    \centering
    \begin{tabular}{@{}lccccc@{}}
        \bottomrule
        \specialrule{0em}{1.5pt}{1.5pt}
        \midrule
        \textbf{Algorithm}                    & \textbf{A}     & \textbf{C}     & \textbf{P}      & \textbf{R}      & \textbf{Avg}     \\
        \midrule
        CAD             & 20.9 $\pm$ 6.9 & 21.3 $\pm$ 9.1 & 31.4 $\pm$ 11.8 & 33.0 $\pm$ 11.7 & 26.6             \\
        IRM       & 41.3 $\pm$ 5.4 & 40.4 $\pm$ 2.7 & 56.6 $\pm$ 5.6  & 60.4 $\pm$ 5.7  & 49.7             \\
        MMD                        & 52.5 $\pm$ 0.2 & 45.3 $\pm$ 0.3 & 66.3 $\pm$ 0.1  & 69.5 $\pm$ 0.6  & 58.4             \\
        ARM            & 50.8 $\pm$ 0.8 & 42.9 $\pm$ 0.5 & 66.0 $\pm$ 0.4  & 67.2 $\pm$ 0.3  & 56.7             \\
        GroupDRO                  & 52.4 $\pm$ 0.7 & 44.7 $\pm$ 1.0 & 67.0 $\pm$ 0.7  & 68.0 $\pm$ 0.7  & 58.0             \\
        VREx                      & 53.4 $\pm$ 0.9 & 45.7 $\pm$ 0.9 & 68.0 $\pm$ 0.1  & 69.6 $\pm$ 0.5  & 59.1             \\
        Bayes-IRM       & 51.4 $\pm$ 0.2 & 46.7 $\pm$ 1.3 & 70.2 $\pm$ 0.6  & 68.9 $\pm$ 1.4  & 59.3             \\
        Mixup            & 52.0 $\pm$ 1.4 & 46.9 $\pm$ 0.7 & 70.2 $\pm$ 0.7  & 71.0 $\pm$ 0.7  & 60.0             \\
        Fishr             & 53.7 $\pm$ 0.5 & 43.7 $\pm$ 0.4 & 67.5 $\pm$ 0.5  & 69.6 $\pm$ 0.1  & 59.1             \\
        SD                        & 54.4 $\pm$ 1.1 & 50.1 $\pm$ 0.4 & 70.3 $\pm$ 0.8  & 73.8$\pm$ 0.7   & \textbf{62.2}    \\
        SagNet                    & 52.2 $\pm$ 1.4 & 47.7 $\pm$ 1.4 & 69.6 $\pm$ 1.1  & 71.1 $\pm$ 0.8  & 60.2             \\
        SelfReg                   & 53.0 $\pm$ 1.2 & 49.2 $\pm$ 0.6 & 70.2 $\pm$ 0.7  & 72.4 $\pm$ 0.7  & 61.2             \\
        Fish                      & 53.9 $\pm$ 0.3 & 48.8 $\pm$ 1.1 & 70.0 $\pm$ 0.2  & 71.9 $\pm$ 0.5  & 61.2             \\
        CORAL                     & 55.4 $\pm$ 0.9 & 48.7 $\pm$ 0.2 & 71.2 $\pm$ 0.6  & 72.2 $\pm$ 0.3  & \underline{61.9} \\
        \midrule
        ERM         & 51.0 $\pm$ 1.6 & 46.8 $\pm$ 1.4 & 68.3 $\pm$ 1.2  & 69.5 $\pm$ 1.5  & 58.9             \\
        PTG-Lite                              & 53.1 $\pm$ 0.1 & 48.4 $\pm$ 0.2 & 70.2 $\pm$ 0.2  & 72.0 $\pm$ 0.4  & 60.9             \\
        ERM-Bayesian & 51.6 $\pm$ 1.0 & 48.4 $\pm$ 0.2 & 66.5 $\pm$ 1.3  & 70.2 $\pm$ 0.4  & 59.2             \\
        PTG                                   & 55.3 $\pm$ 0.5 & 50.8 $\pm$ 0.2 & 69.7 $\pm$ 0.3  & 70.6 $\pm$ 0.4  & \dashuline{61.6} \\
        \hline
        \specialrule{0em}{1.5pt}{1.5pt}
        \bottomrule
    \end{tabular}
\end{table}

\begin{table}[!h]
    \caption{\textbf{TerraIncognita Comparisons}. Out-of-domain classification accuracies(\%) on TerraIncognita are shown. ERM-Bayesian is a BNN \citep{blundell2015weight} trained by ERM. PTG takes ERM-Bayesian as initialization. PTG-Lite takes ERM as initialization. All models are reproduced on DomainBed. We highlight the \textbf{best}, \underline{second} and \dashuline{third} results.}
    \centering
    \begin{tabular}{@{}lccccc@{}}
        \bottomrule
        \specialrule{0em}{1.5pt}{1.5pt}
        \midrule
        \textbf{Algorithm}                    & \textbf{L100}  & \textbf{L38}    & \textbf{L43}   & \textbf{L46}   & \textbf{Avg}     \\
        \midrule
        CAD             & 27.9 $\pm$ 4.7 & 28.8 $\pm$ 10.7 & 31.0 $\pm$ 5.2 & 22.5 $\pm$ 2.8 & 27.5             \\
        IRM       & 37.9 $\pm$ 7.6 & 11.5 $\pm$ 2.4  & 44.2 $\pm$ 2.9 & 35.1 $\pm$ 1.2 & 32.2             \\
        MMD                        & 32.8 $\pm$ 3.0 & 25.7 $\pm$ 1.0  & 47.9 $\pm$ 1.9 & 26.1 $\pm$ 1.8 & 33.1             \\
        ARM            & 40.4 $\pm$ 0.7 & 29.4 $\pm$ 2.4  & 46.9 $\pm$ 0.8 & 29.8 $\pm$ 1.3 & 36.6             \\
        GroupDRO                  & 32.8 $\pm$ 0.7 & 30.2 $\pm$ 2.1  & 48.3 $\pm$ 0.9 & 28.0 $\pm$ 2.1 & 34.8             \\
        VREx                      & 39.2 $\pm$ 4.3 & 32.7 $\pm$ 1.3  & 57.8 $\pm$ 0.8 & 29.7 $\pm$ 3.1 & 37.4             \\
        Bayes-IRM       & 44.0 $\pm$ 2.2 & 29.8 $\pm$ 3.0  & 49.6 $\pm$ 0.6 & 32.0 $\pm$ 2.3 & 38.9             \\
        Mixup            & 49.8 $\pm$ 3.6 & 30.5 $\pm$ 3.9  & 49.9 $\pm$ 0.8 & 31.0 $\pm$ 0.8 & 40.3             \\
        Fishr             & 42.9 $\pm$ 3.9 & 36.6 $\pm$ 0.8  & 48.4 $\pm$ 2.2 & 32.5 $\pm$ 1.0 & 40.1             \\
        SD                        & 40.4 $\pm$ 1.8 & 28.9 $\pm$ 1.7  & 51.7 $\pm$ 0.6 & 33.3 $\pm$ 1.2 & 38.6             \\
        SagNet                    & 42.8 $\pm$ 1.0 & 27.9 $\pm$ 4.4  & 51.1 $\pm$ 1.9 & 35.6 $\pm$ 1.8 & 39.3             \\
        SelfReg                   & 45.1 $\pm$ 2.0 & 30.3 $\pm$ 2.1  & 49.4 $\pm$ 0.4 & 28.0 $\pm$ 1.7 & 38.2             \\
        Fish                      & 42.7 $\pm$ 1.4 & 33.0 $\pm$ 2.9  & 49.1 $\pm$ 0.6 & 31.2 $\pm$ 1.4 & 39.0             \\
        CORAL                     & 45.4 $\pm$ 5.2 & 27.3 $\pm$ 6.3  & 51.4 $\pm$ 2.1 & 30.7 $\pm$ 0.9 & 38.7             \\
        \midrule
        ERM         & 49.5 $\pm$ 3.1 & 32.1 $\pm$ 3.0  & 50.8 $\pm$ 0.1 & 34.2 $\pm$ 0.4 & \dashuline{41.7} \\
        PTG-Lite                              & 53.1 $\pm$ 1.7 & 39.2 $\pm$ 0.8  & 52.1 $\pm$ 0.3 & 35.1 $\pm$ 0.2 & \textbf{44.9}    \\
        ERM-Bayesian & 45.3 $\pm$ 3.5 & 35.3 $\pm$ 1.1  & 49.7 $\pm$ 1.0 & 33.5 $\pm$ 1.6 & 40.9             \\
        PTG                                   & 48.6 $\pm$ 0.8 & 40.7 $\pm$ 0.3  & 52.7 $\pm$ 0.3 & 36.8 $\pm$ 0.4 & \underline{44.7} \\
        \hline
        \specialrule{0em}{1.5pt}{1.5pt}
        \bottomrule
    \end{tabular}
\end{table}

\end{document}